\documentclass[final,3p,times]{elsarticle}              

\usepackage{amssymb}

\usepackage{amssymb,amsfonts,amsmath,amsthm,mathrsfs} 
\usepackage{booktabs}
\usepackage{booktabs}  
\usepackage{makecell}
\usepackage{multirow}  
\usepackage{graphicx}  
\usepackage{threeparttable}  

\usepackage{svg} 

\usepackage{algorithm}
\usepackage{algorithmic}     
\newtheorem{theorem}{Theorem}
\newtheorem{assumption}{Assumption}
\newtheorem{lemma}{Lemma}
\newtheorem{remark}{Remark}

\usepackage{geometry}
\newcommand{\gray}[1]{\textcolor{gray}{#1}}
\usepackage[colorlinks,linkcolor=blue]{hyperref}
\usepackage{longtable}
\usepackage{diagbox} 

\journal{X}

\begin{document}

\begin{frontmatter}



\title{A Triple-Inertial Accelerated Alternating Optimization Method for Deep Learning Training}


\author[1]{Chengcheng Yan}
\ead{ycc956176796@gmail.com}

\author[1]{Jiawei Xu}
\ead{xu_jiawei_2022@163.com}

\author[1]{Qingsong Wang\corref{cor1}}
\ead{nothing2wang@hotmail.com}

\author[1]{Zheng Peng}
\ead{pzheng@xtu.edu.cn}

\address[1]{School of Mathematics and Computational Science, Xiangtan University, Xiangtan, 411105, China}

\cortext[cor1]{Corresponding author}

\begin{abstract}

The stochastic gradient descent (SGD) algorithm has achieved remarkable success in training deep learning models. However, it has several limitations, including susceptibility to vanishing gradients, sensitivity to input data, and a lack of robust theoretical guarantees. In recent years, alternating minimization (AM) methods have emerged as a promising alternative for model training by employing gradient-free approaches to iteratively update model parameters. Despite their potential, these methods often exhibit slow convergence rates. To address this challenge, we propose a novel Triple-Inertial Accelerated Alternating Minimization (TIAM) framework for neural network training. The TIAM approach incorporates a triple-inertial acceleration strategy with a specialized approximation method, facilitating targeted acceleration of different terms in each sub-problem optimization. This integration improves the efficiency of convergence, achieving better performance with fewer iterations. Additionally, we provide a convergence analysis of the TIAM algorithm, including its global convergence properties and convergence rate. Extensive experiments validate the effectiveness of the TIAM method, demonstrating improvements in generalization capability and computational efficiency compared to existing approaches, particularly when applied to the rectified linear unit (ReLU) and its variants.


\end{abstract}







\end{frontmatter}



\section{Introduction}

The stochastic gradient descent (SGD) algorithm is widely recognized as crucial for deep learning model training~\cite{lecun2015deep,lecun1989backpropagation}, as it reduces the computational cost of full-batch gradient calculations, saving both time and memory. In recent years, numerous SGD variants have been proposed, greatly accelerating their adoption in deep learning. However, these methods have inherent limitations. On the one hand, the convergence speed of SGD can be slow and unstable, and its performance is highly sensitive to input data~\cite{novak2018sensitivity}. On the other hand, train performance may degrade due to issues such as vanishing gradients~\cite{hanin2018neural}. Moreover, the theoretical guarantees for applying SGD and its variants to highly non-smooth and non-convex neural network optimization problems remain insufficient.

\begin{figure}
\centering
\includegraphics[width=0.8\linewidth]{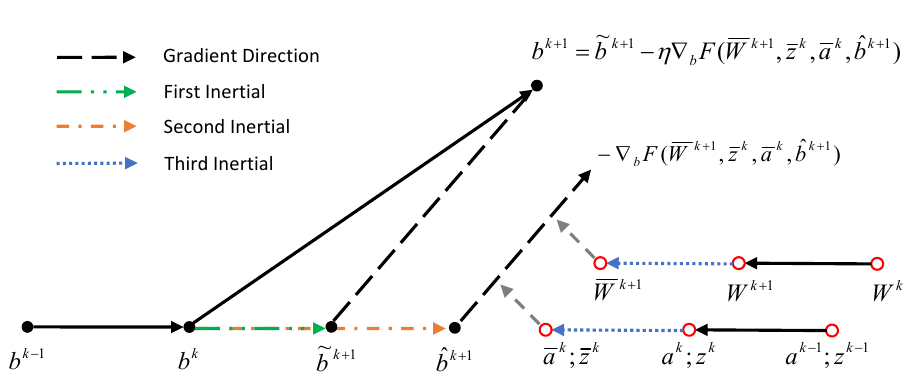}
    \caption{ The figure illustrates the update procedure for the network parameter $b^{k}$ using the triple-inertial acceleration method. $b^{k+1}$ is computed via a conventional gradient descent update within the AM framework. Specifically, $\tilde{b}^{k+1}$ and $\hat{b}^{k+1}$ are derived from the first and second acceleration steps, respectively. Additionally, $\overline{a}^{k}$, $\overline{z}^{k}$, and $\overline{W}^{k+1}$ correspond to the parameter values obtained after the third acceleration step. The proposed framework is highly flexible, allowing seamless incorporation of various inertial acceleration strategies into existing optimization algorithms by effectively leveraging a sufficient amount of historical iterative information. }
\label{fig:TripleAccelarate}
\end{figure}

To overcome the challenges faced by gradient descent algorithms in neural network training, the alternating minimization (AM) method \cite{tseng2001convergence} has emerged as a viable alternative for parameter optimization. The AM method decomposes complex global optimization problems into multiple simple sub-problems, reducing overall complexity by iteratively optimizing the variable of corresponding sub-problems in an alternating fashion. This approach provides a flexible and practical optimization strategy, especially in scenarios involving highly structured or large-scale models, such as compressed sensing (CS)~\cite{sun2016deep} and image processing~\cite{almeida2013deconvolving, 23M1611166}. In this way, model training approaches utilizing AM circumvent the reliance on gradient-based backpropagation, effectively mitigating the vanishing gradient problem. Furthermore, these methods demonstrate better robustness in handling non-smooth regularization terms. Recognizing these advantages, researchers have increasingly focused on the theoretical underpinnings of such methods. For instance, Miguel et al.~\cite{carreira2014distributed} proposed a method for solving constrained optimization problems, applicable to penalty-based approaches by alternating optimization over the parameters and auxiliary coordinates. Based on the block coordinate descent (BCD) method, Lau et al.~\cite{lau2018proximal} proposed a novel algorithm and proved its global convergence, built upon the powerful framework of the Kurdyka-Lojasiewicz (KL) property. Zhang et al.~\cite{zhang2017convergent} prove that the BCD algorithm globally converges to a stationary point with an R-linear convergence rate of order one. For stochastic or mini-batch strategies, Choromanska et al.~\cite{choromanska2019beyond} proposed a novel online (stochastic/mini-batch) AM approach for training deep neural networks, with the first theoretical convergence guarantees for AM in stochastic settings. Furthermore, Tang et al.~\cite{tang2021admmirnn} proposed a distributed paralleled algorithm regarding recurrent neural networks based on the alternating direction method of multipliers (ADMM) framework and proved relevant convergence.

Although the aforementioned methods offer significant advantages in optimizing neural network parameters, they still encounter several challenges:
(1) While the AM method decomposes the original neural network problem into several subproblems, the difficulty of solving certain subproblems can increase computational complexity during the iterative update process, ultimately reducing overall model training efficiency. (2) Neural network model variables often exhibit strong interdependencies or coupling, which may hinder the effective decomposition of subproblems. Therefore, accelerating the solution of subproblems in multi-layer perceptron (MLP) optimization has become a key research focus.

In this work, inspired by previous studies \cite{WANG202366,bolte2014proximal,pock2016inertial}, we extend the triple-inertial acceleration method to solve the MLP problem. 
Specifically, the key contributions of this work are as follows:

\begin{itemize}
    \item \textbf{Triple-Inertial Acceleration Framework:} We propose a simple yet sufficiently general acceleration framework for deep learning model training, employing a triple-inertial acceleration strategy (see Figure \ref{fig:TripleAccelarate} for an illustration) to accelerate parameter updates for each layer. To mitigate the impact of parameter coupling on the optimization process, we propose a method that integrates triple-inertial acceleration techniques with a specialized approximation for the MLP problem. This avoids solving the complexity of the matrix inverse while accelerating the entire iteration process. Additionally, the proposed framework is highly flexible, allowing seamless incorporation of various inertial acceleration strategies into existing optimization algorithms. This significantly broadens the range of solution approaches for MLP problems, enhancing both adaptability and extensibility.
	\item \textbf{Guaranteed Theoretical Convergence:} We provide a convergence analysis of our accelerated optimization algorithm under appropriate assumptions, demonstrating its theoretical effectiveness. It is proven that the proposed algorithm achieves a linear convergence rate.
	\item \textbf{Performance Evaluation and Generalization:} Extensive experiments were conducted on four benchmark datasets, where the proposed method outperforms state-of-the-art algorithms in terms of test accuracy, computational efficiency, and robustness. Furthermore, most previous studies have focused on the rectified linear unit (ReLU) activation function, with limited experimental exploration of its variants. The proposed algorithm demonstrates better generalization capability and robustness for ReLU activation functions and their variants, which have been rarely studied experimentally in prior algorithms.

\end{itemize}

The rest of this paper is organized as follows: Section \ref{sec:related} provides a summary of recent related work. In Section \ref{sec:Algorithm}, we formulate the MLP training problem and introduce the proposed algorithm for training the MLP model. Section \ref{sec:convergence} presents a detailed analysis of the convergence properties of the proposed algorithm. Extensive experimental results on benchmark datasets are presented in Section \ref{sec:Experiments}, and Section \ref{sec:variants} evaluates the performance of variants of the ReLU activation function in our algorithm. Finally, Section \ref{sec:Conclusion} concludes the paper.

\section{Related Work}
\label{sec:related}

Research on deep learning optimization methods can be categorized into two primary approaches: gradient descent-based algorithms and AM methods, as described below:

\subsection{GD Methods}
 The gradient descent (GD) method \cite{mason1999boosting} is one of the most widely used algorithms for solving optimization problems. It is known for its simplicity, ease of implementation, and ability to iteratively approach an optimal solution. SGD~\cite{robbins1951stochastic}, a stochastic variant of GD, has become critically important for optimizing machine learning models, particularly in deep learning. However, it has several limitations. For instance, it is highly sensitive to the learning rate $\alpha$ and prone to becoming trapped in local minima. To mitigate these drawbacks, various variants of the gradient descent algorithm have been developed, broadly categorized into three types: momentum-based methods, adaptive learning rate methods, and regularization techniques. Momentum-based methods, including the Nesterov acceleration~\cite{nesterov1983method}, the Heavy Ball acceleration~\cite{polyak1964some}, and the Optimal Gradient Method~\cite{kim2016optimized}, have been widely used to accelerate neural network training. Adaptive learning rate methods, such as AdaGrad~\cite{duchi2011adaptive}, RMSProp~\cite{tieleman2017divide}, Adam~\cite{kingma2014adam}, AMSGrad~\cite{reddi2018convergence}, and Adabound~\cite{luoadaptive} methods, dynamically adjust the learning rate during training. Regularization techniques, such as $\ell_2$ regularization, help prevent overfitting and improve generalization. Previous studies have shown that the efficiency and convergence of the gradient descent algorithm are crucial for model performance.

\subsection{AM Framework for Deep Learning}
Although SGD methods and their variants are widely used in neural networks, gradient-based backpropagation often suffers from the vanishing gradient problem. Fortunately, researchers have introduced an alternative approach to mitigate this issue using an alternating minimization framework. This method updates neural network parameters alternately, bypassing the backpropagation and thereby avoiding gradient accumulation and vanishing issues.

In recent years, AM algorithms have been widely adopted in deep learning, particularly for non-convex optimization problems or tasks requiring efficient iterative parameter updates \cite{taylor2016training,wang2019admm,gabay1976dual}. In the AM update procedure, the optimization process typically decomposes the original problem into two or more sub-problems, which are then solved alternately through iterative updates. Expanding on this concept, Taylor et al.~\cite{taylor2016training} explored an unconventional training method employing alternating direction methods (scaleADMM) and Bregman iteration to train networks without gradient descent. Inspired by scaleADMM~\cite{taylor2016training}, Wang et al.~\cite{wang2019admm} proposed a novel optimization framework for deep learning via ADMM~\cite{gabay1976dual} (dlADMM) to tackle the lack of global convergence guarantees and slow solution convergence in existing methods. Wang et al.~\cite{wang2022accelerated} extended the work of dlADMM~\cite{wang2019admm} by employing an inequality-constrained formulation to approximate the original problem, which features non-convex equality constraints, thereby facilitating the proof of convergence for the proposed mDLAM algorithm regardless of hyperparameter selection. Wang et al.~\cite{wang2020toward} proposed a novel parallel deep learning ADMM framework (pdADMM) to enable layer parallelism, allowing parameters in each neural network layer to be updated independently and in parallel. To examine the effects of activation functions and batch size on neural network models, Zeng et al.~\cite{zeng2021admm} developed an ADMM-based deep neural network training method that mitigates the saturation issue of sigmoid-type activations. Wang et al.~\cite{wang2024badm} leveraged the ADMM framework to develop a novel data-driven algorithm. The key idea is to partition the training data into batches, which are further subdivided into sub-batches for parameter updates via multi-level aggregation. Furthermore, researchers are increasingly focusing on AM-based neural networks in image processing. Sun et al.~\cite{sun2016deep} proposed a novel deep architecture for optimizing a compressed sensing-based MRI model, ADMM-Net, which is over a data flow graph derived from the iterative procedures in the ADMM algorithm. To reconstruct images from sparsely sampled measurements, Yang et al.~\cite{yang2018admm} proposed a novel deep learning architecture, ADMM-CSNet, which integrates the traditional model-based CS method with a data-driven deep learning approach. These approaches provide a solid foundation for applying the AM framework to neural network training.

\begin{table}[thb]\centering
    \caption{ Notations used in this paper.}
    \label{tab:notations}
    \resizebox{1\textwidth}{!}
    {
    \small
    \begin{tabular}{{cc|cc}}
        \toprule
        Notations & Descriptions & Notations & Descriptions       \\
        \midrule
        $L$ & The total number of layers.    &  $h_l(z_l)$ & The nonlinear activation function in the $l$-th layer. \\ 
        $W_l$ & The weight matrix in the $l$-th layer.   &  $R(z_L;y)$ & The loss function in the $L$-th layer.  \\
        $b_l$ & The bias in the $l$-th layer.              &  $\Omega_l(W_l)$ & The regularization term in the $l$-th layer.\\
        $z_l$ & The output of the linear mapping in the $l$-th layer.   &  $\epsilon$ & The tolerance of the nonlinear mapping.  \\
        $a_l$ & The output of the $l$-th layer.    &  $n_l$ & The number of neurons in the $l$-th layer.  \\
        $x$ & The input matrix of the neural network.     &   $y$ & The predefined label vector.  \\
        ${W}_{l}^{k+1}$  & $W$ at layer $l$ in the $(k+1)$-th iteration.         &  $\mathbf{W}_{l}^{k+1}$ & $\left\{\left\{W_{i}^{k+1}\right\}_{i=1}^{l},\left\{W_{i}^{k}\right\}_{i=l+1}^{L}\right\}$. \\
        ${b}_{l}^{k+1}$  & $b$ at layer $l$ in the $(k+1)$-th iteration.            &  $\mathbf{b}_{l}^{k+1}$ & $\left\{\left\{b_{i}^{k+1}\right\}_{i=1}^{l},\left\{b_{i}^{k}\right\}_{i=l+1}^{L}\right\}$. \\
        ${z}_{l}^{k+1}$  & $z$ at layer $l$ in the $(k+1)$-th iteration.          &  $\mathbf{z}_{l}^{k+1}$ & $\left\{\left\{z_{i}^{k+1}\right\}_{i=1}^{l},\left\{z_{i}^{k}\right\}_{i=l+1}^{L}\right\}$. \\
        ${a}_{l}^{k+1}$  & $a$ at layer $l$ in the $(k+1)$-th iteration.          &  $\mathbf{a}_{l}^{k+1}$ & $\left\{\left\{a_{i}^{k+1}\right\}_{i=1}^{l},\left\{a_{i}^{k}\right\}_{i=l+1}^{L}\right\}$. \\
        \bottomrule
    \end{tabular}
      
    }
\end{table}

\section{Model and Algorithm}
\label{sec:Algorithm}
In this section, we introduce our algorithm in detail. Table \ref{tab:notations} presents the key notations used in this paper. Specifically, $\| \cdot \|$ is defined as the $\ell_2$ norm $\| \cdot \|_2$ for vector inputs, and as the Frobenius norm $\| \cdot \|_F$ for matrix inputs. Section \ref{sec:Problem_foumula} presents the problem formulation in this work. Sections \ref{sec:main_triple_method} and \ref{sec:parameter_update} detail the triple-inertial acceleration process and the update mechanism for each network parameter, respectively. Algorithm \ref{alg:Inertial} presents a detailed example of the triple-inertial acceleration update procedure for $b^{k}$, while Algorithm \ref{alg:our} outlines the proposed TIAM algorithm.

\subsection{Problem Formulation}
\label{sec:Problem_foumula}
In this study, the classical MLP problem consists of $L$ layers, each comprising a linear mapping and a nonlinear activation function. A linear mapping is characterized by a weight matrix $W_l\in\mathbb{R}^{n_l\times n_{l-1}}$, and each layer has a continuous activation function $h_l(\cdot)$ for the nonlinear transformation. In this problem, the output of the \( l \)-th layer, denoted as \( a_l \), is given by $a_l = h_l(z_l)$, where $z_l = W_l a_{l-1} + b_l $, and \( a_{l-1} \in \mathbb{R}^{n_{l-1}} \) represents the input from the \((l-1)\)-th layer. The overall mathematical formulation of the neural network is presented below, where the auxiliary variable $z_l$ is introduced to represent the output of the linear transformation. The general problem is formulated as follows:
\begin{align*}
    &\min_{W_{l},z_{l},a_{l},b_{l}}R(z_{L};y)+\sum_{l=1}^{L}\Omega_{l}(W_{l})\nonumber\\
    &s.t.\quad z_{l}=W_{l}a_{l-1}+b_{l}\quad(l=1,\ldots,L),\\
    &\quad \quad \quad a_{l}=h_{l}(z_{l})\quad(l=1,\ldots,L-1), \nonumber
\end{align*}
where $a_0=x\in\mathbb{R}^d$ is the input of the original neural network, $d$ is the number of feature dimensions, and $y$ denotes the true label vector. The function $R(z_L;y)$ represents the loss function in the $L$-th layer, and $\Omega_l(W_l)\geq0$ is a regularization term in the $l$-th layer. In this work, $R$ and $\Omega_\mathrm{\iota}$ functions are continuous, convex, and proper.

Since most commonly used activation functions are nonlinear, this non-linearity introduces non-convex constraints, which complicate the process of finding an optimal solution to the $z_l$ subproblem. Therefore, in this study, $h_l(z_l)$ is assumed to be quasi-linear. In this manner, the original non-convex constraints can be transformed into inequality constraints, which serve as an infinite approximation of the original optimization problem. We thus reformulate the original problem as follows:
\begin{align}
\label{problem:2}
    &\min_{W_{l},z_{l},a_{l},b_{l}} {F}(\mathbf{W},\mathbf{z},\mathbf{a},\mathbf{b})=R(z_{L};y)+\sum_{l=1}^{L}\Omega_{l}(W_{l}) + \sum_{l=1}^{L} \phi(a_{l-1},W_l,z_l,b_l)\\
    &s.t.\quad h_{l}(z_{l})-\epsilon\leq a_{l}\leq h_{l}(z_{l})+\epsilon\quad(l=1,\cdots,L-1), \nonumber
\end{align}
where the penalty term is given by $\phi(a_{l-1},W_l,z_l,b_l )= \frac{\rho}{2}\left\|z_{l}-W_{l}a_{l-1}-b_{l}\right\|^{2}$. If $\rho\to\infty$ and $\epsilon \to 0$, the above problem approaches the original problem. $\epsilon$ is exerted to project the nonconvex constraints into $\epsilon$-balls. In the above problem, $F(\cdot)$ denotes the objective function, and we define $\mathbf{W}=\{W_l\}_{l=1}^L,\mathbf{z}=\{z_l\}_{l=1}^L,\mathbf{a}=\{a_l\}_{l=1}^{L-1},\mathbf{b}=\{b_l\}_{l=1}^L$. In this paper, we concentrate on solving optimization problem \eqref{problem:2}.

\subsection{Triple-Inertial Acceleration for Alternating Optimization Method}
\label{sec:main_triple_method}

In this work, we primarily employ the AM method to alternately solve different network parameters in the MLP problem in a block-wise manner. This section introduces the triple-inertial alternating optimization method used in our proposed algorithm, which is applied to each block. Here $p_{1},p_{2} \in [0,1)$, $p_{3} \in [0,1)$, and $\eta^k>0$. Firstly, consider the parameters $u$, $v$, and the function $f(u,v)$ as an example. The parameter $u$ is updated according to the following formulation:
\begin{align}
    \label{algo:acc1}
    \tilde{u}_{k+1}=u_{k}+p_{1}(u_{k}-u_{k-1}), \nonumber \\
    \hat{u}_{k+1}=u_{k}+p_{2}(u_{k}-u_{k-1}), \nonumber \\
    u_{k+1} = \textbf{Operator}(\tilde{u}_{k+1}-\eta_{1}^{k}{\nabla}_{u}f(\hat{u}_{k+1},\overline{v}_{k})), \\
    \overline{u}_{k+1}=u_{k+1}+p_{3}(u_{k+1}-u_{k}). \nonumber 
\end{align}

Secondly, after obtaining the extrapolated variable $\overline{u}_{k+1}$, we use it to update another parameter, $v$. In this work, the $\mathbf{Operator}$ symbol denotes the $\arg\min$ update method for the parameter.
\begin{align}
    \label{algo:acc2}
    \tilde{v}_{k+1}=v_{k}+p_{1}(v_{k}-v_{k-1}),  \nonumber \\
    \hat{v}_{k+1}=v_{k}+p_{2}(v_{k}-v_{k-1}),  \nonumber \\
    v_{k+1} = \textbf{Operator}(\tilde{v}_{k+1}-\eta_{2}^{k}{\nabla}_{v}f(\overline{u}_{k+1},\hat{v}_{k+1})), \\
    \overline{v}_{k+1}=v_{k+1}+p_{3}(v_{k+1}-v_{k}).  \nonumber 
\end{align}

Finally, we extend the above two-block update scheme to a multi-block setting in our algorithm. Let all neural network parameters $\mathbf{W},\mathbf{b},\mathbf{z},\mathbf{a}$ be updated in the $k$-th iteration following the above alternating update process. Algorithm \ref{alg:Inertial} outlines the framework for updating parameter $b$. $\eta_1^{k}$ and $\eta_2^{k}$ are learned using the backtracking algorithm~\cite{wang2019admm}.

\begin{algorithm}[H]
\caption{The Triple-Inertial Accelerate Framework to Update $b_{l}^{k}$}\label{alg:Inertial}
\begin{algorithmic}[1]
\label{algo:block}
\STATE $\mathbf{Input:~} \mathbf{W},\mathbf{b},\mathbf{z},\mathbf{a},\overline{W}_{l}^{k+1},\overline{z}_l^{k},\overline{a}_{l-1}^{k}$. $\mathbf{Output:~}b_{l}^{k+1},\overline{b}_{l}^{k+1}$.
\STATE $\tilde{b}_{l}^{k+1}=b_{l}^{k}+p_{1}(b_{l}^{k}-{b}_{l}^{k-1})$   \quad \gray{ \text{(\# First Accelerate)}   }
\STATE $ \hat{b}_{l}^{k+1}=b_{l}^{k}+p_{2}(b_{l}^{k}-{b}_{l}^{k-1})$    \quad \gray{ \text{(\# Second Accelerate)}   }
\STATE $ b_{l}^{k+1} = \arg\min\limits_{b_l} (\tilde{b}_{l}^{k+1}-\eta_{1}^{k}{\nabla}_{b} {F}(\overline{W}_{l}^{k+1},\overline{z}_l^{k},\overline{a}_{l-1}^{k},\hat{b}_{l}^{k+1}))$ 
\IF{${F}(\mathbf{W}_{l-1}^{k+1},\mathbf{z}_{l-1}^{k+1},\mathbf{a}_{l-1}^{k+1},\mathbf{b}_{l}^{k+1}) \geq {F}(\mathbf{W}_{l-1}^{k+1},\mathbf{z}_{l-1}^{k+1},\mathbf{a}_{l-1}^{k+1},\mathbf{b}_{l-1}^{k+1})$}
    \STATE $ b_{l}^{k+1} = \arg\min\limits_{b_l} ({b}_{l}^{k}-\eta_{1}^{k}{\nabla}_{b} {F}( {W}_{l}^{k+1},{z}_l^{k}, {a}_{l-1}^{k},{b}_{l}^{k}))$  \quad \gray{ \text{(\# If $b_{l}^{k+1}$ increase $F(\cdot)$ compared $b_{l}^{k}$)}}
\ENDIF

\STATE $ \overline{b}_{l}^{k+1}= b_{l}^{k+1}+p_{3}( b_{l}^{k+1}-b_{l}^{k})$    \quad \gray{ \text{(\# Third Accelerate)}   }
\STATE \textbf{Output} $b_{l}^{k+1},\overline{b}_{l}^{k+1}$
\end{algorithmic}
\end{algorithm}

\begin{remark}
The distinction between the TIAM and mDLAM algorithms and special cases.
\begin{itemize}
    \item The key difference between the TIAM and mDLAM \cite{wang2022accelerated} algorithms is the placement of acceleration steps. Most existing AM approaches, including mDLAM, used for solving MLP problems either omit acceleration strategies or apply a single, identical inertial acceleration before subproblem updates. However, these studies have largely overlooked valuable historical iteration data both before and after subproblem updates. Effectively utilizing this information can further improve algorithmic efficiency. The proposed method applies distinct inertial accelerations for the previous step term and gradient term respectively before subproblem updates, followed by an additional extrapolation operation on the remaining parameters after subproblem updates. This allows the algorithm to promptly incorporate the latest historical information throughout the optimization process, continuously refining and correcting the current iteration. Notably, the first two acceleration steps in TIAM primarily focus on the current subproblem update, while the third acceleration is specifically designed to propagate extrapolation effects to other subproblem iterations.
    \item If $p_1 = p_2$ is set to the Nesterov acceleration coefficient~\cite{nesterov1983method}, $p_3 = 0$, and the other hyperparameters without adaptive adjustments, TIAM reduces to mDLAM.
\end{itemize}

\end{remark}

In Algorithm \ref{alg:our}, $b_{l}^{k+1}$ and $\overline{b}_{l}^{k+1}$ are the updated values of $b \in \mathbf{b}=\{b_l\}_{l=1}^L$, computed using Algorithm \ref{alg:Inertial}. $b_{l}^{k-1}$ represents the parameter from the $(k-1)$-th iteration in the $l$-th layer, $W_{l}^{k},z_l^{k}$, and $a_{l}^{k}$ are elements of $\mathbf{W}=\{W_l\}_{l=1}^{L},\mathbf{z}=\{z_l\}_{l=1}^L$, and $\mathbf{a}=\{a_l\}_{l=1}^{L-1}$ in the $k$-th epoch. The update sequence for all parameters follows: $\mathbf{W} \rightarrow \mathbf{b} \rightarrow \mathbf{z} \rightarrow \mathbf{a}$. The proposed algorithm ensures a decrease in the objective function $F(\cdot)$. Taking the parameter $b_l^{k+1}$ as an example in Algorithm \ref{alg:Inertial}, if the value of $F(\cdot)$ increases after an update, all accelerated variables will revert to their non-accelerated states for a repeated update. This means that $b_{l}^{k+1}$ will be updated again by setting $\tilde{b}_l^{k+1} = {b}_l^{k}, \hat{b}_l^{k+1} = {b}_l^{k}, \overline{W}_{l}^{k+1} = {W}_{l}^{k+1}, \overline{z}_l^{k} = {z}_l^{k}, \overline{a}_{l-1}^{k} = {a}_{l-1}^{k}, \overline{b}_{l}^{k} = {b}_{l}^{k}$, ensuring a decrease in $F(\cdot)$. This process, reflected in line 6 of Algorithm \ref{alg:Inertial}, is also applied to other network parameters $\mathbf{W},\mathbf{z},\mathbf{a}$. Figure \ref{fig:TripleAccelarate} illustrates the triple-inertial acceleration procedure.

\begin{algorithm}[H]
\caption{The proposed TIAM algorithm}
\label{alg:our}
\begin{algorithmic}[1]
\STATE \textbf{Input:} $y$, $a_{0}=x$. \textbf{Output:} $\textbf{W},\textbf{b},\textbf{z},\textbf{a}$.
\STATE Initialize, $p_1$, $p_2$, $p_3$, $k=0$, $K=200$
    \FOR{$k=1$ to $K$}
        \FOR{$l=1$ to $L$}
            \STATE Accelerate and update $W_{l}^{k+1},\overline{W}_{l}^{k+1}$ with (\ref{eq:argminW}).
            \STATE Accelerate and update $b_{l}^{k+1},\overline{b}_{l}^{k+1}$ with (\ref{eq:argminb}).
            \IF{$l = L$}
                \STATE Accelerate and update $z_{L}^{k+1},\overline{z}_{L}^{k+1}$ with (\ref{eq:zL_argmin}).
            \ELSE
                \STATE Accelerate and update $z_{l}^{k+1},\overline{z}_{l}^{k+1}$ with (\ref{eq:zl_argmin}).
            \ENDIF
            \STATE Accelerate and update $a_{l}^{k+1},\overline{a}_{l}^{k+1}$ with (\ref{eq:argmina}).
        \ENDFOR
    \ENDFOR
\STATE \textbf{until} Satisfy terminate condition
\STATE \textbf{Output} $\textbf{W},\textbf{b},\textbf{z},\textbf{a}$
\end{algorithmic}
\end{algorithm}

\begin{remark}
Our proposed method exhibits the following characteristics and advantages:
\begin{itemize}
    \item Although previous AM algorithms have made significant progress in solving MLP problems, most algorithms primarily employ single-step acceleration. In contrast, the proposed algorithm integrates a triple-inertial acceleration strategy into the alternating update process of network parameters at each layer, as illustrated in Figure \ref{fig:TripleAccelarate}. This strategy substantially accelerates convergence and swiftly attains optimal test accuracy. Additionally, we present a convergence analysis of the proposed algorithm, covering both global convergence and its rate.
    \item To avoid computationally expensive matrix inversion in coupled terms, we introduce a specialized approximation approach that approximates the original problem using a backtracking algorithm. This approximation method differs from the quadratic approximation in mDLAM \cite{wang2022accelerated}, as it applies distinct inertial acceleration operations to different parameters.
    \item To the best of our knowledge, most existing AM methods for solving MLP problems primarily utilize the ReLU activation function in experiments, with limited exploration of its variants. In this study, we perform a comprehensive comparison of various ReLU variants to demonstrate the better performance of our algorithm on ReLU-type activation functions.
\end{itemize}
\end{remark}

\subsection{Network Parameter Updates}
\label{sec:parameter_update}

In this section, we provide a detailed exposition of these concepts, including the implementation of the triple-inertial acceleration strategy, the specialized approximation method, and the application of the alternating minimization approach for parameter updates, as outlined in Algorithm \ref{alg:Inertial}.

\subsubsection{Update $W_{l}$}
The variables $W_l$ for $l=1, \cdots, L$ are updated as follows:
\begin{align}
    \label{eq:w}
    W_l^{k+1}\leftarrow\arg\min_{W_l}\phi\left( {a}_{l-1}^{k+1},{W}_l, {z}_l^k, {b}_l^k\right)+\Omega_l(W_l),
\end{align}
since $W_{l}$ and $a_{l-1}$ are coupled in $\phi(\cdot)$, solving for $W_l$ requires inverting $a_{l-1}^{k+1}$, which is computationally expensive. Inspired by the dlADMM algorithm, we introduce $P_l^{k+1}\left(W_l;\theta_l^{k+1}\right)$ as an approximation of $\phi$ around $W_l$. Under the triple-inertial acceleration method, the transformed formula is given by
\begin{align}
P_{l}^{k+1}\left(W_{l};\theta_{l}^{k+1}\right)&= \phi\left(\overline{a}_{l-1}^{k+1},\tilde{W}_{l}^{k+1},\overline{z}_{l}^{k},\overline{b}_l^k\right)+\left(\nabla_{{W}_{l}}\phi\left(\overline{a}_{l-1}^{k+1},\hat{W}_{l}^{k+1},\overline{z}_{l}^{k},\overline{b}_l^k\right)\right)^{T} \left(W_{l}-\tilde{W}_{l}^{k+1}\right)+\frac{\theta_{l}^{k+1}}{2}\|W_{l}-\tilde{W}_{l}^{k+1}\|^{2}. 
\end{align}

Let $\tilde{W}_{l}^{k+1}$ and $\hat{W}_{l}^{k+1}$ represent the previous iteration term and the traditional gradient term within the $\arg\min$ procedure, respectively, each obtained through a distinct inertial acceleration technique. Similarly, $\overline{a}_{l-1}^{k+1},\overline{z}_l^k$, and $\overline{b}_l^k$ represent the outputs of the third inertial acceleration for the corresponding parameters. Here $\theta_l^{k+1}>0$ is a scalar parameter that can be determined using the backtracking algorithm~\cite{wang2019admm} to satisfy the following condition: 
\begin{align}
\label{eq:leqwphi}
P_l^{k+1}\left(W_l^{k+1};\theta_l^{k+1}\right)\geqslant\phi\Big( {a}_{l-1}^{k+1},W_l^{k+1}, {z}_l^k, {b}_l^k\Big),
\end{align}
different to minimizing (\ref{eq:w}), we choose to minimize the following:
\begin{align}
\label{eq:argminW}
W_l^{k+1}\leftarrow\arg\min_{W_l}P_l^{k+1}\left(W_l;\theta_l^{k+1}\right)+\Omega_l(W_l)  ,
\end{align}
where $\Omega_l(W_l)$ means common regularization terms like $\ell_1$ or $\ell_2$ regularization lead to closed-form solutions.

\subsubsection{Update $b_{l}$}

The variables $b_{l}$ for $l=1, \cdots, L$ are updated as follows:
\begin{align}
    \label{eq:b}
    b_{l}^{k+1}\leftarrow\underset{b_{l}}{\operatorname*{argmin}} \phi\left( {a}_{1-1}^{k+1}, {W}_{l}^{k+1}, {z}_{l}^{k},b_{l}\right),
\end{align}
similar to update $W_l$, we define $U_l^{k+1}\left(b_l;\xi_l^{k+1}\right)$ as an approximation of $\phi$ at $b_l$ as follows:

\begin{align}
U_{l}^{k+1}\left(b_{l};\xi_{l}^{k+1}\right)&=\phi\left(\overline{a}_{l-1}^{k+1},\overline{W}_{l}^{k+1},\overline{z}_{l}^{k},\tilde{b}_{l}^{k+1}\right)+\left(\nabla_{{b}_{l}}\phi\left(\overline{a}_{l-1}^{k+1},\overline{W}_{l}^{k+1},\overline{z}_{l}^{k},\hat{b}_{l}^{k+1}\right)\right)^{T}  \left(b_{l}-\tilde{b}_{l}^{k+1}\right)+\frac{\xi_{l}^{k+1}}{2}\|b_{l}-\tilde{b}_{l}^{k+1}\|^{2}, 
\end{align}
where $\xi_l^{k+1}>0$ is a scalar parameter, which can be chosen by the backtracking algorithm \cite{wang2019admm} to meet the following condition:
\begin{align}
U_l^{k+1}\left(b_l^{k+1};\xi_l^{k+1}\right)\geqslant\phi\Big( {a}_{l-1}^{k+1}, {W}_{l}^{k+1}, {z}_{l}^{k}, b_{l}^{k+1}\Big).
\end{align}
The original problem can be reformulated as follows:
\begin{align}
\label{eq:argminb}
b_l^{k+1}\leftarrow\arg\min_{b_l}U_l^{k+1}\left(b_l;\xi_l^{k+1}\right),
\end{align}
so the solution can be obtained as follows since it is convex and has a closed-form solution:
\begin{align}
b_l^{k+1}\leftarrow \tilde{b}_l^{k+1}-\nabla_{{b}_l}\phi\left(\overline{a}_{l-1}^{k+1},\overline{W}_{l}^{k+1},\overline{z}_{l}^{k},\hat{b}_{l}^{k+1}\right)/ \xi_l^{k+1}   .
\end{align}

\subsubsection{Update $z_{l}$}
The variables $z_l$ for $l=1, \cdots, L$ are updated as follows:
\begin{align}
    \label{eq:z}
    &z_{l}^{k+1}\leftarrow\arg\min_{z_{l}} \phi\left( {a}_{l-1}^{k+1}, {W}_{l}^{k+1},z_{l}, {b}_l^{k+1}\right), \nonumber\\
    &s.t. \quad h_l(z_l)-\epsilon\leqslant a_l\leqslant h_l(z_l)+\epsilon \quad(l<L)  ,
\end{align}
and for $z_L^{k+1}$
\begin{align}
    \label{eq:zL}
    z_{L}^{k+1}\leftarrow\arg\min_{z_{L}} \phi( {a}_{L-1}^{k+1}, {W}_{L}^{k+1},z_{L}, {b}_{L}^{k+1})+R(z_{L};y).
\end{align}
Similar to updating $W_l$, we define $V_l^{k+1}(z_l)$ as follows:
\begin{align}
V_{l}^{k+1}(z_{l})=&\phi\left(\overline{a}_{l-1}^{k+1},\overline{W}_{l}^{k+1},\tilde{z}_{l}^{k+1},\overline{b}_l^{k+1}\right)+\left(\nabla_{{z}_{l}}\phi\left(\overline{a}_{l-1}^{k+1},\overline{W}_{l}^{k+1},\hat{z}_{l}^{k+1},\overline{b}_l^{k+1}\right)\right)^{T}\left(z_{l}-\tilde{z}_{l}^{k+1}\right)  +\frac\rho2\|z_l-\tilde{z}_l^{k+1}\|^2  , 
\end{align}
so the original problem can be reformulated to solve the following problems:
\begin{align}
\label{eq:zl_argmin}
&z_{l}^{k+1} \leftarrow \arg\min_{z_{l}} V_{l}^{k+1}(z_{l}), \quad \text{s.t.} \quad h_{l}(z_{l}) - \epsilon \leqslant a_{l} \leqslant h_{l}(z_{l}) + \epsilon \quad (l < L). \\
\label{eq:zL_argmin}
&z_{L}^{k+1} \leftarrow \arg\min_{z_{L}} V_{L}^{k+1}(z_{L}) + R(z_{L}; y).
\end{align}

As for $z_l (l=1,\cdots,l-1)$, the solution can be transformed as
\begin{align}
z_l^{k+1}\leftarrow\min\left(\max\left(B_1^{k+1},\tilde{z}_l^{k+1}-\nabla_{{z}_l}\phi\left(\overline{a}_{l-1}^{k+1},\overline{W}_{l}^{k+1},\hat{z}_{l}^{k+1},\overline{b}_l^{k+1}\right)/\rho\right),B_2^{k+1}\right),
\end{align}
where $B_1^{k+1}$ and $B_2^{k+1}$ denote the lower bound and the upper bound of the set ${z_l}$. Equation \eqref{eq:zL_argmin} can be efficiently solved using the fast iterative soft-thresholding algorithm (FISTA)~\cite{beck2009fast}.

\subsubsection{Update $a_{l}$}
The variables $a_l$ for $l=1, \cdots, L-1$ are updated as follows:
\begin{align}
    \label{eq:a}
    &a_l^{k+1}\leftarrow\arg\min_{a_l}\phi(a_l, {W}_{l+1}^k, {z}_{l+1}^k, {b}_{l+1}^{k}),   \nonumber \\ 
    &s.t. \quad h_l({z}_l^{k+1})-\epsilon\leqslant a_l\leqslant h_l({z}_l^{k+1})+\epsilon. 
\end{align}
Similar to updating $W_l^{k+1},Q_l^{k+1}(a_l;\tau_l^{k+1})$ is defined as
\begin{align}
Q_{l}^{k+1}\left(a_{l};\tau_{l}^{k+1}\right)=&\phi\left(\tilde{a}_{l}^{k+1},\overline{W}_{l+1}^{k},\overline{z}_{l+1}^{k},\overline{b}_{l+1}^{k}\right)+\left(\nabla_{{a}_{l}}\phi\left(\hat{a}_{l}^{k+1},\overline{W}_{l+1}^{k},\overline{z}_{l+1}^{k},\overline{b}_{l+1}^{k}\right)\right)^{T}\left(a_{l}-\tilde{a}_{l}^{k+1}\right) +\frac{\tau_l^{k+1}}2\|a_l-\tilde{a}_l^{k+1}\|^2,  
\end{align}
and then we can solve the following problem instead:
\begin{align}
\label{eq:argmina}
a_l^{k+1}\leftarrow\arg\min_{a_l}Q_l^{k+1}(a_l;\tau_l^{k+1}),\:s.t.\:h_l(\overline{z}_l^{k+1})-\epsilon\leqslant a_l\leqslant h_l(\overline{z}_l^{k+1})+\epsilon,
\end{align}
where $\tau_l^{k+1}>0$ is a scalar parameter, similar to the update procedure of the aforementioned parameters, which the backtracking algorithm can choose to meet the following condition:
\begin{align}
Q_l^{k+1}(a_l^{k+1};\tau_l^{k+1})\geqslant\phi\Big(a_l^{k+1}, {W}_{l+1}^k, {z}_{l+1}^k, {b}_{l+1}^{k}\Big).
\end{align}
Thus, the solution can be obtained by
\begin{align}
a_l^{k+1}\leftarrow\:\min\left(\max\left(h_l(\overline{z}_l^{k+1})-\epsilon,\tilde{a}_l^{k+1}-\nabla_{{a}_l}\phi\left(\hat{a}_{l}^{k+1},\overline{W}_{l+1}^{k},\overline{z}_{l+1}^{k},\overline{b}_{l+1}^{k}\right)/\tau_l^{k+1}\right),h_l(\overline{z}_l^{k+1})+\epsilon\right).
\end{align}

\section{Convergence Analysis}
\label{sec:convergence}

This section discusses the convergence of the TIAM algorithm. Detailed proofs are provided in the Appendix, and the following mild assumption is required for the convergence analysis of the proposed TIAM algorithm:

\begin{assumption}
\label{assu:1}
$F(\mathbf{W}, \mathbf{z}, \mathbf{a}, \mathbf{b})$ is coercive over the domain $\{ (\mathbf{W}, \mathbf{z}, \mathbf{a}, \mathbf{b}) \mid h_l(z_l) - \epsilon \leq a_l \leq h_l(z_l) + \epsilon, \, (l = 1, \ldots, L - 1) \}$.
\end{assumption}

Assumption \ref{assu:1} is mild and holds for common loss functions~\cite{wang2019admm}, such as the cross-entropy loss.

\subsection{Convergence Properties}

This section outlines the convergence properties of the proposed algorithm. Following standard procedures in optimization theory, this analysis primarily focuses on the following lemma:

\begin{lemma}
\label{lemma:F}
There exist $\beta_l,\delta_l,\gamma_l,\zeta_l>0$ for $k\in\mathbb{N}$ such that it holds that
\begin{align}
    \label{eq:Conver_ineq1}
    F(\mathbf{W}_{l-1}^{k+1}, \mathbf{z}_{l-1}^{k+1}, \mathbf{a}_{l-1}^{k+1}, \mathbf{b}_{l-1}^{k+1}) - F(\mathbf{W}_{l}^{k+1}, \mathbf{z}_{l-1}^{k+1}, \mathbf{a}_{l-1}^{k+1}, \mathbf{b}_{l-1}^{k+1}) &\geq \frac{\beta_{l}^{k+1}}{2} \left\| W_{l}^{k+1} - W_{l}^{k} \right\|^{2}, \\
    \label{eq:Conver_ineq2}
    F(\mathbf{W}_{l}^{k+1}, \mathbf{z}_{l-1}^{k+1}, \mathbf{a}_{l-1}^{k+1}, \mathbf{b}_{l-1}^{k+1}) - F(\mathbf{W}_{l}^{k+1}, \mathbf{z}_{l}^{k+1}, \mathbf{a}_{l-1}^{k+1}, \mathbf{b}_{l-1}^{k+1}) &\geq \frac{\delta_{l}^{k+1}}{2} \left\| z_{l}^{k+1} - z_{l}^{k} \right\|^{2}, \\
    \label{eq:Conver_ineq3}
    F(\mathbf{W}_{l}^{k+1}, \mathbf{z}_{l}^{k+1}, \mathbf{a}_{l-1}^{k+1}, \mathbf{b}_{l-1}^{k+1}) - F(\mathbf{W}_{l}^{k+1}, \mathbf{z}_{l}^{k+1}, \mathbf{a}_{l}^{k+1}, \mathbf{b}_{l-1}^{k+1}) &\geq \frac{\gamma_{l}^{k+1}}{2} \left\| a_{l}^{k+1} - a_{l}^{k} \right\|^{2}, \\
    \label{eq:Conver_ineq4}
    F(\mathbf{W}_{l}^{k+1}, \mathbf{z}_{l-1}^{k+1}, \mathbf{a}_{l-1}^{k+1}, \mathbf{b}_{l-1}^{k+1}) - F(\mathbf{W}_{l}^{k+1}, \mathbf{z}_{l-1}^{k+1}, \mathbf{a}_{l-1}^{k+1}, \mathbf{b}_{l}^{k+1}) &\geq \frac{\zeta_{l}^{k+1}}{2} \left\| b_{l}^{k+1} - b_{l}^{k} \right\|^{2}.
\end{align}
\end{lemma}
Lemma \ref{lemma:F} states that the objective function $F(\cdot)$ decreases with the update of all parameters. Under Assumption \ref{assu:1} and Lemma \ref{lemma:F}, the following theorem presents the convergence properties:

\begin{lemma}
    \label{lemma:decrease}
    (Objective Decrease) In Algorithm \ref{algo:block}, it holds that for any $k\in\mathbb{N},F\left(\mathbf{W}^{k},\mathbf{z}^{k},\mathbf{a}^{k},\mathbf{b}^{k}\right)\geqslant F\left(\mathbf{W}^{k+1},\mathbf{z}^{k+1},\mathbf{a}^{k+1},\mathbf{b}^{k+1}\right).$ Besides, $F$ is convergent. Specifically, $F\left(\mathbf{W}^{k},\mathbf{z}^{k},\mathbf{a}^{k},\mathbf{b}^{k}\right)\to F^{*}$ as $k\to\infty$, where $F^{*}$ is the convergent value of $F.$
\end{lemma}

This lemma ensures both the decrease and the convergence of the objective function, providing a theoretical foundation for the stability and efficiency of the proposed algorithm.

\begin{lemma}
\label{lemma:bounder}
(Bounded Objective and Variables) In Algorithm \ref{alg:our}, it holds that for any $k\in\mathbb{N}$, 
\begin{itemize} 
\item[(a)] $F(\mathbf{W}^k, \mathbf{z}^k, \mathbf{a}^k, \mathbf{b}^k)$ is upper bounded. Moreover, $\lim_{k \to \infty} \| \mathbf{W}^{k+1} - \mathbf{W}^{k} \| = 0, \lim_{k \to \infty} \| \mathbf{z}^{k+1} - \mathbf{z}^{k} \| = 0 ,\lim_{k \to \infty}  \| \mathbf{a}^{k+1} - \mathbf{a}^{k} \| = 0$ ~and~ $\lim_{k \to \infty}  \| \mathbf{b}^{k+1} - \mathbf{b}^{k} \| = 0$.

\item[(b)] $(\mathbf{W}^k, \mathbf{z}^k, \mathbf{a}^k, \mathbf{b}^k)$ is bounded. Specifically, there exist scalars $M_\mathbf{W}, M_\mathbf{z}, M_\mathbf{a}$ and $M_\mathbf{b}$ such that $\|\mathbf{W}^k\| \leq M_\mathbf{W}, \|\mathbf{z}^k\| \leq M_\mathbf{z}, \|\mathbf{a}^k\| \leq M_\mathbf{a}$ and $\|\mathbf{b}^k\| \leq M_\mathbf{b}$. 
\end{itemize}
\end{lemma}

This lemma guarantees that both the objective function and all variables remain bounded within the proposed algorithm. Furthermore, the difference between corresponding variables in consecutive iterations, such as $\mathbf{W}^{k+1}$ and $\mathbf{W}^{k}$, converges to zero.

\begin{lemma}
    \label{lemma:subgradient}
    (Subgradient Bound) In Algorithm \ref{alg:our}, it exist that: $C_{1}= \max\left( M , M +\theta_{1}^{k+1}, M +\theta_{2}^{k+1},\cdots,  M +\theta_{L}^{k+1}\right)$, and $g_1^{k+1}\in\partial_{\mathbf{W}^{k+1}}F$ such that for any $k\in\mathbb{N}$,
    \begin{align*}
        &\|g_1^{k+1}\|\leqslant C_1\Big(\|\mathbf{W}^{k+1}-\mathbf{W}^k\|+\|\mathbf{z}^{k+1}-\mathbf{z}^k\|+ \|\mathbf{a}^{k+1}-\mathbf{a}^k\| +\|\mathbf{b}^{k+1}-\mathbf{b}^k\|+\|\mathbf{W}^k-\mathbf{W}^{k-1}\|  \nonumber + \|\mathbf{b}^k-\mathbf{b}^{k-1}\| + \|\mathbf{z}^k-\mathbf{z}^{k-1}\| \Big) .
    \end{align*}

\end{lemma}

Lemma~\ref{lemma:subgradient} indicates that the subgradient of the objective is bounded in terms of its variables. This implies that the subgradient converges to zero, thereby establishing convergence to a stationary point. The proof of all the above lemmas is provided in Appendix \hyperref[lemma:final_result]{B}.

\subsection{ Convergence of the Proposed Algorithm}
Based on the preceding lemmas, including objective decrease, bounded objective and variable, and subgradient bound, we derive the following convergence theorem:

\begin{theorem}
\label{theory:conver}
(Convergence under the Triple-Inertial Acceleration Method) Suppose that Assumption \ref{assu:1} holds. Under the safeguarding step of the triple-inertial acceleration method, for any $\rho>0$, $p_{1},p_{2} \in [0,1)$, $p_{3} \in [0,1)$ and $\epsilon>0$, Algorithm \ref{alg:our} converges to a stationary point $\mathbf{W}^{*}$. That is $0\in\partial F_{\mathbf{W}^*}(\mathbf{W}^*,\mathbf{z}^*,\mathbf{a}^*,\mathbf{b}^*).$
\end{theorem}

This theorem guarantees that the convergence always holds no matter how $\mathbf{W}$ is initialized, and whatever $\rho$, $\epsilon$, $p_1$, $p_2$ and $p_3$ are chosen. Based on Theorem \ref{theory:conver}, we can derive the convergence rate of the proposed algorithm.

\begin{theorem}
\label{theory:rate}
(Convergence Rate) In our algorithm, If $F$ is a lower semicontinuous convex subanalysis function, the convergence rate of $F(\mathbf{W}^k, \mathbf{z}^k, \mathbf{a}^k, \mathbf{b}^k)$ is Linear Convergence Rate. If $F$ is locally strongly convex, then for any $\rho$, there exist $\epsilon > 0$, $k_1 \in \mathbb{N}$ and $0 < C_{9} < 1$ such that it holds for $k > k_1$ that
\begin{align*}
F\left( \mathbf{W}^{k+1}, \mathbf{z}^{k+1}, \mathbf{a}^{k+1} , \mathbf{b}^{k+1} \right) - F^* \leq C_9 \left( F\left( \mathbf{W}^{k-1}, \mathbf{z}^{k-1}, \mathbf{a}^{k-1} , \mathbf{b}^{k-1}\right) - F^* \right).
\end{align*}
\end{theorem} 

This theorem shows that our algorithm converges linearly for sufficiently large iterations. All these proofs are given in Appendix \hyperref[converge:final_result]{C}.

\section{Experiments}
\label{sec:Experiments}

Following the experimental settings of the existing algorithm, we constructed a fully connected neural network as our base model. The network consists of three layers, each containing 100 hidden units, with the ReLU activation function~\cite{nair2010rectified} applied to each neuron. The selected loss function is cross-entropy with softmax, and the total number of training epochs is set to 200. To mitigate the effects of random initialization of weights and biases, we conducted experiments using 10 different random seeds ranging from 0 to 9. The final evaluation results are presented as the mean and standard deviation of the corresponding epochs, and all the comparison algorithms follow the same settings. The experiments were implemented in Python 3.8 with Torch 1.9~\cite{paszke2017automatic} and conducted on a Windows 10 computer equipped with an Intel(R) Core(TM) i7-12700 2.10 GHz CPU and 16GB of memory. The implementation will be publicly accessible at \url{https://github.com/yancc103/TIAM} once this manuscript is accepted.

\subsection{Comparison Method and Experimental Settings}

In our work, we employ GD methods and their variants, along with dlADMM \cite{wang2019admm} and mDLAM \cite{wang2022accelerated}, which are considered state-of-the-art optimizers, for comparative analysis. The models are trained on a full-batch dataset, with all parameters selected to achieve the highest training accuracy. The baseline methods are described as follows:

\textbf{GD}: GD is one of the most widely used optimization algorithms in deep learning, where model parameters are iteratively updated in the direction of the negative gradient. Its convergence properties have been extensively studied in the literature.

\textbf{Adagrad}: The Adaptive Gradient method is a variant of gradient descent that adjusts the learning rate for each parameter individually, allowing it to adapt dynamically during training. This adaptive mechanism improves optimization efficiency, particularly for sparse data.

\textbf{Adadelta}: The Adaptive Learning Rate method extends Adagrad by mitigating its sensitivity to hyperparameters. It modifies the update rule for the learning rate to improve performance and stability over time, eliminating the need for a manually predefined learning rate.

\textbf{Adam}: The Adaptive Momentum Estimation is one of the most widely adopted optimization algorithms in deep learning, known for its strong generalization performance. It refines gradient updates by incorporating estimates of the first and second moments, leading to faster convergence and reduced bias.

\textbf{AM methods}: The dlADMM and mDLAM algorithms are two representative AM methods designed for general MLP optimization. These approaches reformulate neural network parameter updates as subproblem solutions, with theoretical guarantees on their convergence properties.

Table \ref{tab:parameters} summarizes the hyperparameters utilized across all methods. In both our proposed method and the mDLAM algorithm, $\rho$ controls the quadratic terms in the MLP problem. The learning rate $\alpha$ is applied in the comparison methods, except for the AM methods. In the dlADMM algorithm, $\rho$ regulates the linear constraint. Additionally, the tolerance hyperparameter $\epsilon$ is adaptively adjusted as follows: $\epsilon^{k+1} = \max(\epsilon^k / 2, 0.0001)$, with the initial value $\epsilon^0 = 100$. This adaptive scheme relaxes the inequality constraints in the early stages and progressively tightens them as training advances. When $\epsilon^k$ is large, the constraints become easier to satisfy.

\begin{table}[!thb]\centering
    \caption{The hyperparameter configuration of four benchmark datasets.}
    \label{tab:parameters}
    \resizebox{1\textwidth}{!}
    {
    \begin{tabular}{*{6}{c}}
        \toprule
        Method &Hyper-parameters  &Cora   &Amazon Photo   & Citeseer  &Coauthor CS \\
        \midrule
        TIAM (ours) & $\rho$ & $1 \times 10^{-4}$  &$4 \times 10^{-4}$ & $3\times 10^{-5}$ & $3\times 10^{-5}$ \\
        mDLAM & $\rho$ & $1 \times 10^{-3}$  &$1 \times 10^{-3}$ & $5\times 10^{-3}$ & $1\times 10^{-4}$ \\
        dlADMM & $\rho$ & $1 \times 10^{-6}$  &$1 \times 10^{-6}$ & $1\times 10^{-6}$ & $1\times 10^{-6}$ \\
        GD & $\alpha$ & 0.01  &0.01 & 0.01 & $5\times 10^{-3}$ \\
        Adadelta & $\alpha$ & 0.01  &$1 \times 10^{-3}$ & 0.01 & 0.05   \\
        Adagrad & $\alpha$ & $5 \times 10^{-3}$  &$1 \times 10^{-3}$ & 0.01 & $5\times 10^{-3}$    \\ 
        Adam & $\alpha$ & $1 \times 10^{-3}$  &$1 \times 10^{-3}$ & $1\times 10^{-3}$ & $1\times 10^{-3}$ \\
        \bottomrule
    \end{tabular}
    }
\end{table}

With respect to the acceleration hyperparameters settings of our proposed algorithm, we implemented adaptive adjustments for $p_1, p_2, p_3$, and $\rho$. Specifically\footnote{Here, the settings of $p_1$ and $p_2$ are inspired by \cite{bolte2014proximal}, and experimental results indicate that nonlinear decay is more effective than linear decay strategy for $p_3$. So determining a better acceleration coefficient remains an important direction for future research.}, $p_1 = \frac{k-1}{k+2} \times \check{p}_1$, $p_2 = \frac{k-1}{k+2} \times \check{p}_2$ and $p_3 = (1 - \frac{k}{K})^{1.25} \times \check{p}_3 $, where $\check{p}_1 = \check{p}_2 = 1$ to simplify the choice of hyperparameters, $\check{p}_3 = 0.55$, and $K$ denotes the total number of iterations. It is worth noting that all three inertial acceleration coefficients can be different, they are defined as functions of the iteration number $k$. Moreover, different coefficient settings may lead to varying test performance. Additionally, the parameter $\rho$ was adaptively adjusted in each iteration $k$, the adjustment rule is defined as $\rho = \mathrm{max}(1.2 \times \rho, 0.001)$ if the condition satisfies:  $\small( train\_cost^{k} \geq train\_cost^{k-1} \mathrm{~and~} train\_cost^{k-1} \geq train\_cost^{k-2} \small)$.

\subsection{Datasets}
In this section, we evaluate the effectiveness of our algorithm using the following four benchmark datasets\footnote{Dataset available at \url{https://github.com/shchur/gnn-benchmark/}}. The partitioning of train and test sets for all datasets follows the same scheme as the mDLAM method~\cite{wang2022accelerated}, with the full batch dataset utilized for model training.

\textbf{Cora}~\cite{sen2008collective}: The Cora dataset includes 2708 machine learning research papers, which are categorized into 7 classes. The citation network contains 5429 edges, with a dictionary of 1433 unique words.

\textbf{Amazon Photo}~\cite{shchur2018pitfalls}: The Amazon Photo dataset is a graph where nodes represent photo-related products, such as cameras and tripods, while edges signify co-purchase relationships, indicating products often bought together. It comprises 7487 nodes and 238162 edges, with each node described by a 745-dimensional feature vector extracted from product data. The nodes are grouped into 8 classes. 

\textbf{Citeseer}~\cite{sen2008collective}: The Citeseer dataset consists of 3327 scientific papers sourced from the Tagged.com social network. It is classified into 6 categories, with 4732 citation links and a vocabulary of 3703 unique words (i.e. feature).  

\textbf{Coauthor CS}~\cite{shchur2018pitfalls}: Derived from the Microsoft Academic Graph for the KDD Cup 2016 competition, this dataset consists of authors as nodes, with edges indicating co-authorship. The keywords from each author's papers serve as node features, and class labels correspond to their primary research areas. It comprises 18333 nodes and 81,894 edges, organized into 15 distinct classes. A 6805-dimensional feature vector characterizes each node.

\subsection{Performance and Efficiency}

\begin{figure}
\centering
\includegraphics[width=0.83\linewidth]{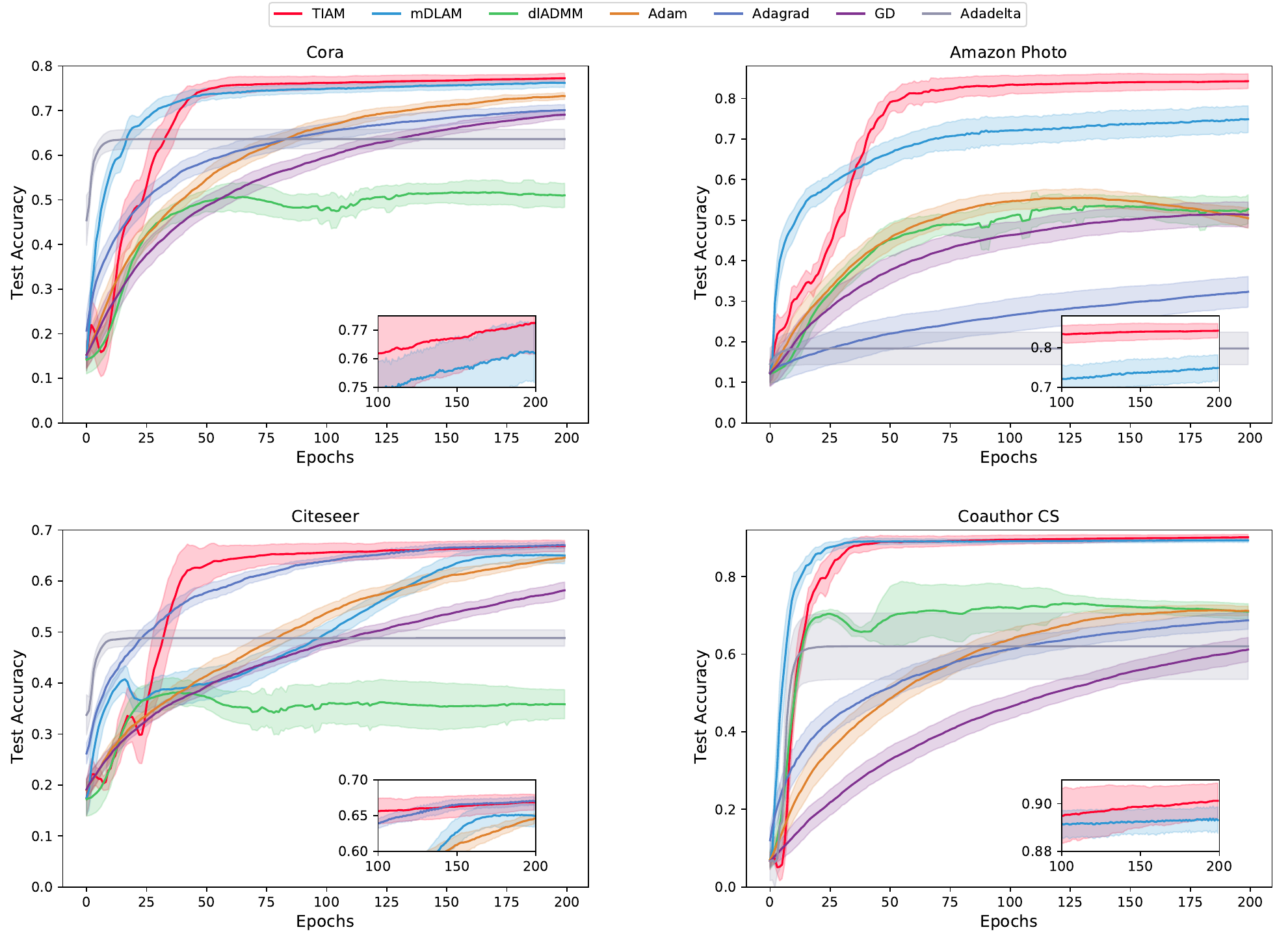}
\caption{The curve illustrates the mean and standard deviation of test accuracy for all methods over 10 runs. The proposed algorithm consistently outperforms all other comparison methods for the four datasets.}
\label{fig:relu_acc}
\end{figure}

\begin{figure}
\centering
\includegraphics[width=0.8\linewidth]{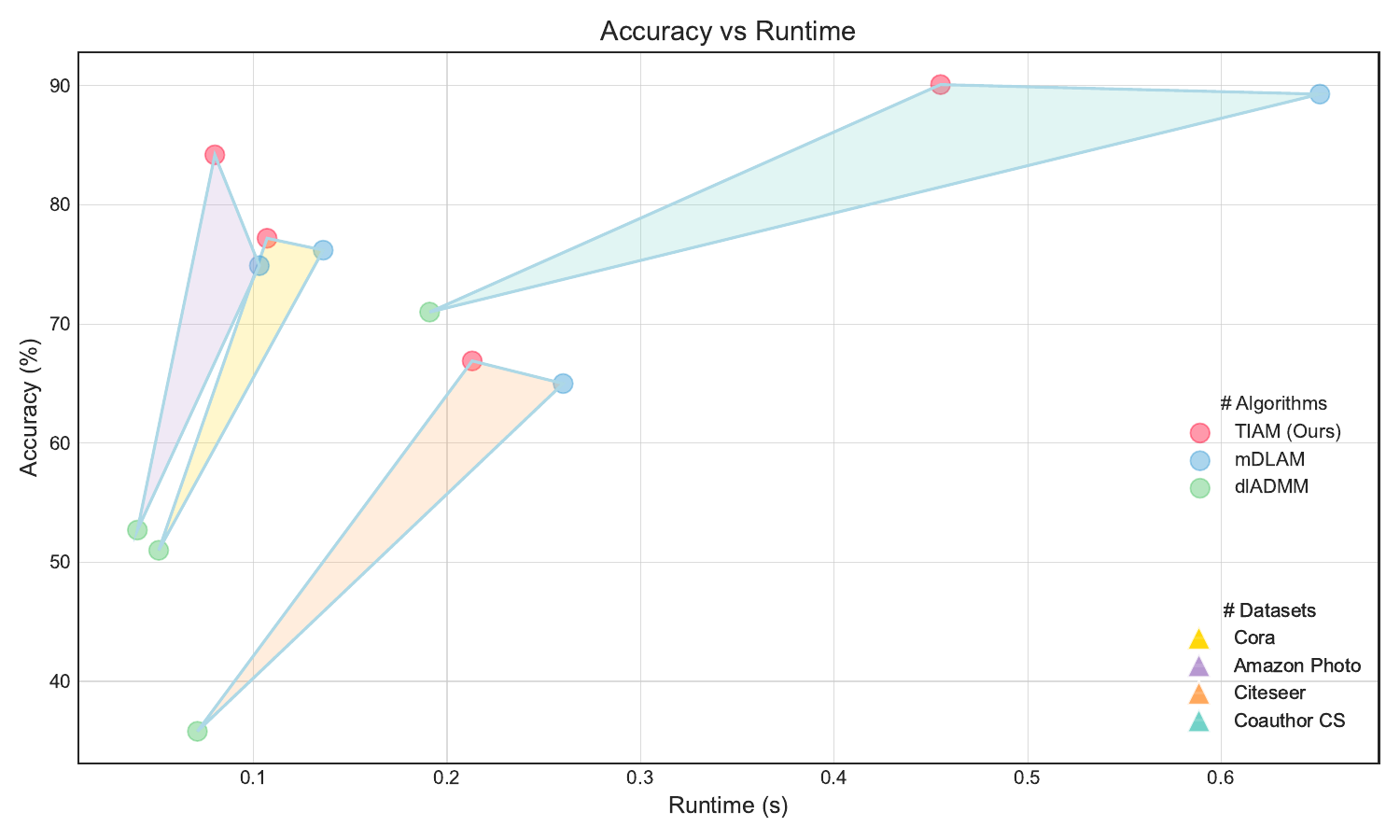}
\caption{The relationship between test accuracy and runtime for the AM methods in the four datasets.}
\label{fig:acc_vs_runtime}
\end{figure}

\begin{figure}
\centering
\includegraphics[width=0.9\linewidth]{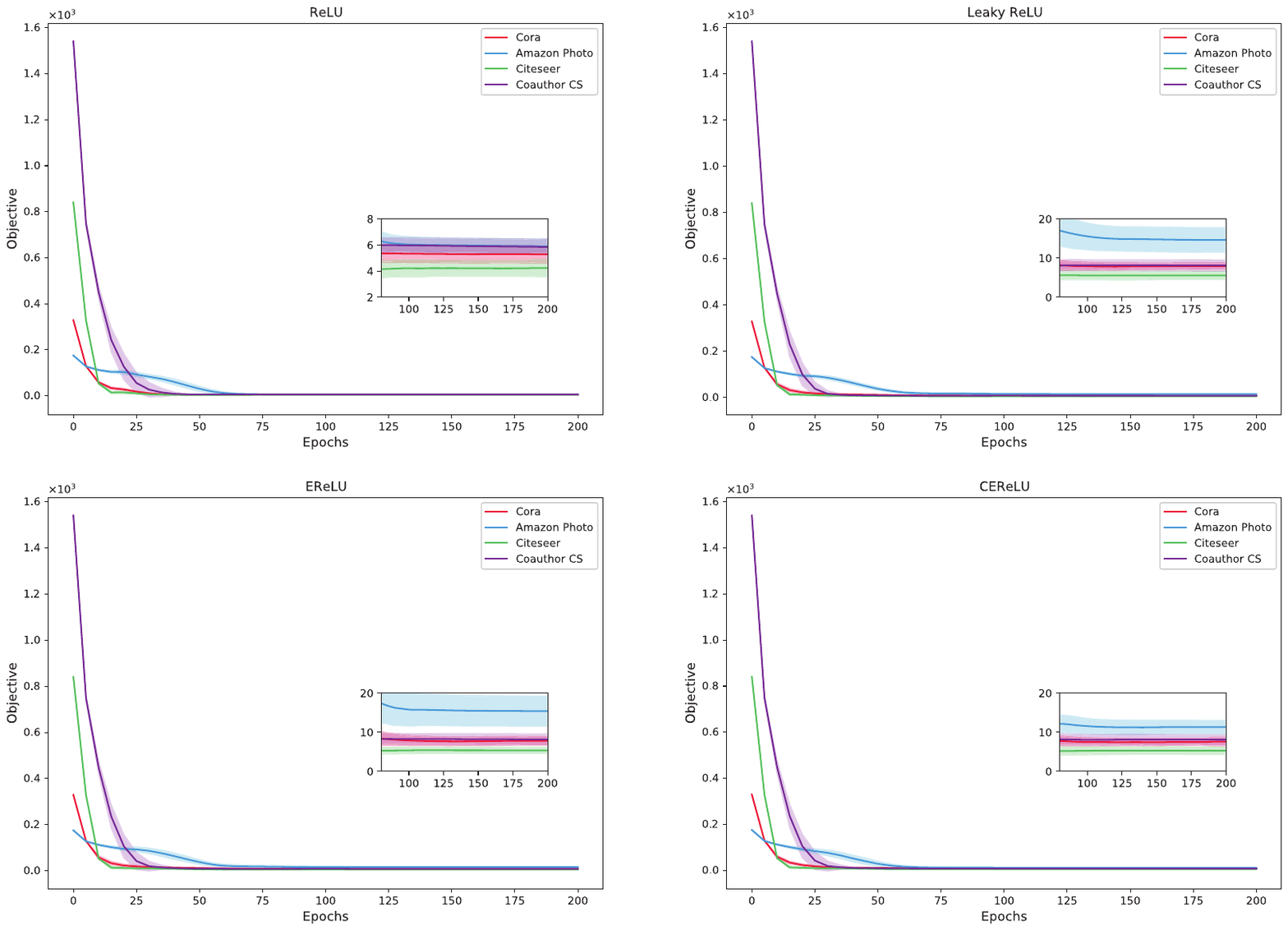}
\caption{The curve illustrates the mean and standard deviation of the objective function values achieved by the proposed method over 10 runs, utilizing the ReLU activation function and its variants.}
\label{fig:objective_curve}
\end{figure}

In this section, we analyze the performance and efficiency of our method, focusing on test accuracy, the decrease in the objective function, and the runtime.

Figure \ref{fig:relu_acc} illustrates the test accuracy curves of all comparison algorithms utilizing the ReLU activation function in the experiment. To eliminate differences caused by the random initialization of neural network parameters, we fixed random seeds (i.e. 0 to 9) for all compared algorithms over 10 runs, ensuring consistent and reproducible experimental results. Across four datasets, our proposed algorithm consistently achieves the highest test accuracy under the same epoch conditions. Although our algorithm converges more slowly in the early stages compared to the mDLAM algorithm, it achieves higher accuracy in the later stages, with most datasets surpassing the mDLAM algorithm in accuracy after 35 epochs.

Furthermore, Figure \ref{fig:objective_curve} illustrates the overall decreasing trend of the objective function values for the optimization problem \eqref{problem:2}, which aligns with the theoretical guarantee of objective reduction established in Lemma \ref{lemma:decrease}. This shows that within the alternating optimization framework, the objective function exhibits a general downward trend, thereby confirming the consistency between the experimental results and the theoretical expectations of the proposed method.

Regarding runtime in the comparison experiment, Figure \ref{fig:acc_vs_runtime} illustrates the relationship between test accuracy and average runtime per epoch for the AM method presented in this paper. In our experiment, we first summed the runtime per epoch over 10 runs and then averaged the results to obtain the final runtime. From the graph, it can be observed that our method achieves the highest test accuracy within a relatively short runtime for all datasets. Notably, the comparison experiment reveals significant differences among the three algorithms on the Coauthor CS dataset, highlighting that our proposed method performs well on larger datasets.

\subsection{Sensitivity Analysis}

Although our proposed algorithm demonstrated significant advantages on four datasets in previous experiments, it is also crucial to investigate the impact of various factors on the algorithm's performance, referred to as sensitivity analysis. In this section, we conduct a comprehensive sensitivity analysis of our algorithm from four perspectives: an ablation study on the acceleration coefficient, the effect of the acceleration coefficient $p_3$, the influence of the hyperparameter $\rho$, and the impact of varying the number of neurons. These analyses offer deeper insights into the algorithm's sensitivity to different factors.

\subsubsection{Ablation Study for the TIAM Method}

In this section, we compare the test accuracy performance on certain accelerated coefficient while fixing other coefficients. As demonstrated in Table \ref{tab:ablation}, $T_{1,2,3}$ represents an acceleration coefficient flag (i.e. $p_1,p_2,p_3$) for the triple-inertial acceleration method, whereas the baseline refers to the case where all acceleration coefficients in the proposed algorithm are set to zero. The full model represents the complete TIAM algorithm, which incorporates the triple-inertial acceleration strategy.

In the ablation experiment, we coupled the first and second acceleration strategies to maintain consistency with the previous acceleration coefficient settings. As shown in Table \ref{tab:ablation}, for most datasets, the $T_{1,2}$ acceleration strategy yields better performance in the early stages. In contrast, the $T_{3}$ acceleration strategy, although initially achieving lower accuracy, contributes more to the final test accuracy in later stages. This finding underscores the critical role of the variable values obtained from the third acceleration in the later stages of the update process. This variable is closely related to the three remaining neural network parameters, excluding the current update parameter. Leveraging the variables from the third acceleration enables the algorithm to identify the optimal solution throughout the optimization process, thereby enhancing its overall accuracy. Thus, $p_3$ plays a crucial role in our algorithm. In the following sections, we will analyze the impact of different coefficient $p_3$ settings on the performance of the proposed algorithm.

\begin{table*}[thb]\centering
    \caption{Ablation study with test accuracy for our method over 10 runs.}
    \label{tab:ablation}
    \resizebox{1\textwidth}{!}
    {
    \begin{tabular}{c|*{5}{c}}
        \toprule
        \multicolumn{6}{c}{Cora} \\
        \midrule
        \diagbox{Module}{Epoch} & 40 & 80  &120  &160  &200 \\
        \midrule
        Baseline  & $0.699\pm0.029$ & $0.720\pm0.024$ & $0.721\pm0.024$ & $0.722\pm0.024$ & $0.722\pm0.024$  \\
        w/ ${T_{1,2}}$ & $0.718\pm0.017$ & $0.726\pm0.015$ & $0.731\pm0.012$ & $0.731\pm0.012$ & $0.732\pm0.011$ \\
        w/ ${T_{3}}$ & $0.522\pm0.034$ & $0.650\pm0.025$ & $0.708\pm0.026$ & $0.752\pm0.019$ & $0.770\pm0.012$ \\
        \textbf{Full Model} & $0.703\pm0.036$ & $0.760\pm0.016$ & $0.764\pm0.014$ & $0.768\pm0.011$ & $0.772\pm0.011$ \\
        
        \midrule
        \multicolumn{6}{c}{Amazon Photo} \\
        \midrule
        Baseline  & $0.607\pm0.035$ & $0.654\pm0.041$ & $0.673\pm0.037$ & $0.683\pm0.035$ & $0.689\pm0.035$ \\
        w/ ${T_{1,2}}$  & $0.657\pm0.055$ & $0.690\pm0.048$ & $0.696\pm0.047$ & $0.698\pm0.048$ & $0.700\pm0.048$ \\
        w/ ${T_{3}}$ & $0.368\pm0.033$ & $0.535\pm0.026$ & $0.627\pm0.039$ & $0.703\pm0.051$ & $0.760\pm0.045$ \\
        \textbf{Full Model} & $0.668\pm0.042$ & $0.828\pm0.025$ & $0.836\pm0.022$ & $0.840\pm0.020$ & $0.842\pm0.018$  \\
        \midrule
        
        \multicolumn{6}{c}{Citeseer}  \\
        \midrule
        Baseline  & $0.594\pm0.029$ & $0.634\pm0.029$ & $0.636\pm0.027$ & $0.637\pm0.027$ & $0.637\pm0.027$ \\
        w/ ${T_{1,2}}$ & $0.616\pm0.030$ & $0.638\pm0.027$ & $0.643\pm0.028$ & $0.647\pm0.023$ & $0.648\pm0.022$ \\
        w/ ${T_{3}}$ & $0.435\pm0.037$ & $0.592\pm0.042$ & $0.651\pm0.032$ & $0.661\pm0.016$ & $0.666\pm0.013$ \\
        \textbf{Full Model} & $0.599\pm0.058$ & $0.652\pm0.019$ & $0.658\pm0.015$ & $0.664\pm0.014$ & $0.669\pm0.011$ \\
        \midrule
        
        \multicolumn{6}{c}{Coauthor CS}  \\
        \midrule
        Baseline  & $0.857\pm0.013$ & $0.870\pm0.014$ & $0.885\pm0.011$ & $0.891\pm0.009$ & $0.891\pm0.009$ \\
        w/ ${T_{1,2}}$ & $0.850\pm0.021$ & $0.872\pm0.013$ & $0.879\pm0.010$ & $0.884\pm0.008$ & $0.891\pm0.006$ \\
        w/ ${T_{3}}$ & $0.594\pm0.044$ & $0.786\pm0.014$ & $0.838\pm0.012$ & $0.868\pm0.010$ & $0.883\pm0.010$ \\
        \textbf{Full Model} & $0.883\pm0.019$ & $0.892\pm0.013$ & $0.896\pm0.011$ & $0.899\pm0.008$ & $0.901\pm0.007$ \\
        \bottomrule
    \end{tabular}
    }
\end{table*}

\subsubsection{Effect of Acceleration Coefficient $p_3$}

As shown in Table \ref{tab:ablation}, $p_3$ plays a crucial role in the triple-inertial acceleration process, as observed in the ablation study of the TIAM method. To further investigate its impact, we conducted comparative experiments to evaluate test performance under different initial values of $p_3$. Table \ref{tab:p_3} shows that when $p_3$ is selected from the set $\{0.45, 0.55, 0.65\}$ in increasing order, test accuracy improves with increasing $p_3$. The performance on each dataset reaches its peak when $p_3$ is set to 0.55. However, increasing $p_3$ beyond this point does not consistently enhance performance. Beyond this peak, test accuracy for all datasets begins to decline, particularly for the Citeseer dataset, where the algorithm’s final performance drops sharply from 0.669 to 0.300 when $p_3$ is set to 0.65.

\begin{table}[thb]\centering
    \caption{ The impact of $p_3$ on four datasets' mean and standard deviation of test accuracy over 10 runs.}
    \label{tab:p_3}
    \resizebox{1\textwidth}{!}
    {
    \begin{tabular}{c|*{5}{c}}
        \toprule
        \multicolumn{6}{c}{Cora} \\
        \midrule
        \diagbox{$p_3$}{Epoch} & 40 & 80  &120  &160  &200 \\
        \midrule
        0.45 & $0.704\pm0.029$ & $0.738\pm0.022$ & $0.750\pm0.019$ & $0.757\pm0.017$ & $0.761\pm0.015$ \\
        0.55 & $0.703\pm0.036$ & $0.760\pm0.016$ & $0.764\pm0.014$ & $0.768\pm0.011$ & $0.772\pm0.011$ \\
        0.65 & $0.375\pm0.083$ & $0.641\pm0.072$ & $0.689\pm0.051$ & $0.701\pm0.043$ & $0.708\pm0.041$  \\
        \midrule
        \multicolumn{6}{c}{Amazon Photo} \\
        \midrule
        0.45 & $0.769\pm0.030$ & $0.803\pm0.032$ & $0.813\pm0.031$ & $0.818\pm0.032$ & $0.822\pm0.029$ \\
        0.55 & $0.668\pm0.042$ & $0.828\pm0.025$ & $0.836\pm0.022$ & $0.840\pm0.020$ & $0.842\pm0.018$ \\
        0.65 &  $0.441\pm0.074$ & $0.715\pm0.023$ & $0.770\pm0.027$ & $0.784\pm0.025$ & $0.790\pm0.026$  \\
        \midrule
        
        \multicolumn{6}{c}{Citeseer}  \\
        \midrule
        0.45 & $0.637\pm0.043$ & $0.651\pm0.034$ & $0.662\pm0.024$ & $0.670\pm0.019$ & $0.673\pm0.015$ \\
        0.55 & $0.599\pm0.058$ & $0.652\pm0.019$ & $0.658\pm0.015$ & $0.664\pm0.014$ & $0.669\pm0.011$ \\
        0.65 & $0.249\pm0.037$ & $0.274\pm0.052$ & $0.291\pm0.050$ & $0.297\pm0.056$ & $0.300\pm0.056$   \\
        \midrule
        
        \multicolumn{6}{c}{Coauthor CS}  \\
        \midrule
        0.45 & $0.866\pm0.015$ & $0.877\pm0.016$ & $0.884\pm0.013$ & $0.890\pm0.009$ & $0.893\pm0.007$ \\
        0.55 & $0.883\pm0.019$ & $0.892\pm0.013$ & $0.896\pm0.011$ & $0.899\pm0.008$ & $0.901\pm0.007$ \\
        0.65 &  $0.721\pm0.054$ & $0.813\pm0.041$ & $0.822\pm0.039$ & $0.827\pm0.040$ & $0.829\pm0.039$  \\
        \bottomrule
    \end{tabular}
    }
\end{table}

\subsubsection{Sensitivity to Hyper-parameters $\rho$}

As emphasized in previous studies, $\rho$ is a critical hyperparameter in the algorithm. Table \ref{tab:rho_exp} shows that when hyper-parameter $\rho$ is selected from the set $\{1\times10^{-3},1\times10^{-4},1\times10^{-5},1\times10^{-6}\}$, the optimal value of $\rho$ varies for different datasets. Similar to the existing AM method, the selection of $\rho$ significantly influences the algorithm’s performance. This is because $\rho$ plays a pivotal role in the alternating optimization process by determining the tightness of the penalty constraint $ \frac{\rho}{2}\left|z_{l}-W_{l}a_{l-1}-b_{l}\right|^{2}$, which in turn affects the convergence of the algorithm. In this analysis, we observe that when $\rho$ is set to $1\times10^{-6}$, the proposed algorithm performs poorly across all datasets. However, when $\rho$ is within the range of $1\times10^{-5}$ to $1\times10^{-4}$, the algorithm generally exhibits better performance. 

\subsubsection{Test Accuracy of Different Hidden Units}

Finally, Table \ref{tab:hidden_exp} presents the test accuracy of our method for different numbers of hidden units. The number of neurons plays a vital role in the convergence process of neural networks. Selecting an appropriate number of neurons can significantly improve the generalization ability and test accuracy of neural network models. In our experiment, we varied the number of neurons by selecting values from the set $\{100, 200, 400, 600\}$, while keeping all other parameters fixed, to rigorously assess the sensitivity of our algorithm to the number of neurons. As shown in Table \ref{tab:hidden_exp}, the accuracy of our algorithm does not always improve as the number of neurons increases. This phenomenon aligns with a well-known conclusion in deep learning. Increasing the number of neurons indefinitely does not lead to a continuous improvement in algorithm accuracy. Therefore, selecting an appropriate number of neurons is essential.

\begin{table}[thb]\centering
    \caption{ The impact of $\rho$ on four datasets' mean and standard deviation of test accuracy over 10 runs.}
    \label{tab:rho_exp}
    \resizebox{1\textwidth}{!}
    {
    \begin{tabular}{c|*{5}{c}}
        \toprule
        \multicolumn{6}{c}{Cora} \\
        \midrule
        \diagbox{$\rho$}{Epoch} & 40 & 80  &120  &160  &200 \\
        \midrule
        $1\times 10^{-3}$ &  $0.512\pm0.030$ & $0.703\pm0.026$ & $0.728\pm0.021$ & $0.737\pm0.018$ & $0.745\pm0.018$  \\
        $1\times 10^{-4}$ &  $0.703\pm0.036$ & $0.760\pm0.016$ & $0.764\pm0.014$ & $0.768\pm0.011$ & $0.772\pm0.011$   \\
        $1\times 10^{-5}$ &  $0.612\pm0.071$ & $0.660\pm0.049$ & $0.675\pm0.044$ & $0.693\pm0.041$ & $0.708\pm0.036$  \\
        $1\times 10^{-6}$ &  $0.339\pm0.142$ & $0.380\pm0.133$ & $0.400\pm0.137$ & $0.425\pm0.133$ & $0.466\pm0.122$ \\
        
        \midrule
        \multicolumn{6}{c}{Amazon Photo} \\
        \midrule
        $1\times 10^{-3}$ &   $0.609\pm0.051$ & $0.794\pm0.037$ & $0.815\pm0.033$ & $0.820\pm0.031$ & $0.825\pm0.030$  \\
        $1\times 10^{-4}$ &   $0.636\pm0.078$ & $0.736\pm0.065$ & $0.753\pm0.059$ & $0.759\pm0.057$ & $0.763\pm0.059$   \\
        $1\times 10^{-5}$ &   $0.376\pm0.049$ & $0.443\pm0.074$ & $0.458\pm0.079$ & $0.465\pm0.086$ & $0.470\pm0.095$    \\
        $1\times 10^{-6}$ &   $0.214\pm0.074$ & $0.216\pm0.072$ & $0.218\pm0.069$ & $0.216\pm0.060$ & $0.217\pm0.054$  \\
        \midrule
        
        \multicolumn{6}{c}{Citeseer}  \\
        \midrule
        $1\times 10^{-3}$ &   $0.322\pm0.025$ & $0.391\pm0.046$ & $0.416\pm0.049$ & $0.433\pm0.050$ & $0.449\pm0.050$   \\
        $1\times 10^{-4}$ &   $0.381\pm0.036$ & $0.479\pm0.047$ & $0.513\pm0.049$ & $0.534\pm0.053$ & $0.545\pm0.055$ \\
        $1\times 10^{-5}$ &   $0.599\pm0.031$ & $0.627\pm0.024$ & $0.632\pm0.025$ & $0.643\pm0.022$ & $0.649\pm0.020$     \\
        $1\times 10^{-6}$ &   $0.391\pm0.112$ & $0.433\pm0.113$ & $0.442\pm0.107$ & $0.475\pm0.095$ & $0.496\pm0.087$   \\
        \midrule
        
        \multicolumn{6}{c}{Coauthor CS}  \\
        \midrule
        $1\times 10^{-3}$ &   $0.618\pm0.032$ & $0.745\pm0.023$ & $0.821\pm0.017$ & $0.838\pm0.012$ & $0.841\pm0.011$   \\
        $1\times 10^{-4}$ &   $0.770\pm0.034$ & $0.855\pm0.015$ & $0.859\pm0.012$ & $0.860\pm0.013$ & $0.863\pm0.012$   \\
        $1\times 10^{-5}$ &   $0.795\pm0.075$ & $0.824\pm0.058$ & $0.846\pm0.044$ & $0.863\pm0.030$ & $0.876\pm0.024$   \\
        $1\times 10^{-6}$ &   $0.566\pm0.040$ & $0.581\pm0.042$ & $0.601\pm0.040$ & $0.613\pm0.042$ & $0.646\pm0.042$  \\
        \bottomrule
    \end{tabular}
    }
\end{table}

\begin{table}[thb]\centering
    \caption{ The impact of the hidden units on four datasets' mean and standard deviation of test accuracy over 10 runs.}
    \label{tab:hidden_exp}
    \resizebox{1\textwidth}{!}
    {
    \begin{tabular}{c|*{5}{c}}
        \toprule
        \multicolumn{6}{c}{Cora} \\
        \midrule
        \diagbox{Hidden Units}{Epoch} & 40 & 80  &120  &160  &200 \\
        \midrule
        $100$ &  $0.703\pm0.036$ & $0.760\pm0.016$ & $0.764\pm0.014$ & $0.768\pm0.011$ & $0.772\pm0.011$  \\
        $200$ &  $0.667\pm0.098$ & $0.738\pm0.039$ & $0.745\pm0.034$ & $0.750\pm0.032$ & $0.754\pm0.030$   \\
        $400$ &  $0.632\pm0.047$ & $0.737\pm0.039$ & $0.737\pm0.038$ & $0.736\pm0.035$ & $0.737\pm0.034$  \\
        $600$ &  $0.625\pm0.039$ & $0.691\pm0.024$ & $0.691\pm0.024$ & $0.690\pm0.023$ & $0.692\pm0.024$    \\
        \midrule
        \multicolumn{6}{c}{Amazon Photo} \\
        \midrule
        $100$ &   $0.668\pm0.042$ & $0.828\pm0.025$ & $0.836\pm0.022$ & $0.840\pm0.020$ & $0.842\pm0.018$  \\
        $200$ &   $0.707\pm0.023$ & $0.825\pm0.024$ & $0.831\pm0.021$ & $0.836\pm0.021$ & $0.839\pm0.019$  \\
        $400$ &   $0.652\pm0.033$ & $0.803\pm0.024$ & $0.807\pm0.027$ & $0.806\pm0.026$ & $0.805\pm0.026$ \\
        $600$ &   $0.670\pm0.046$ & $0.839\pm0.021$ & $0.839\pm0.022$ & $0.839\pm0.021$ & $0.842\pm0.023$ \\
        \midrule
        
        \multicolumn{6}{c}{Citeseer}  \\
        \midrule
        $100$ &  $0.599\pm0.058$ & $0.652\pm0.019$ & $0.658\pm0.015$ & $0.664\pm0.014$ & $0.669\pm0.011$   \\
        $200$ &   $0.618\pm0.042$ & $0.648\pm0.024$ & $0.654\pm0.022$ & $0.656\pm0.022$ & $0.662\pm0.020$  \\
        $400$ &   $0.585\pm0.056$ & $0.601\pm0.056$ & $0.601\pm0.056$ & $0.602\pm0.055$ & $0.602\pm0.056$  \\
        $600$ &   $0.611\pm0.062$ & $0.633\pm0.044$ & $0.630\pm0.042$ & $0.629\pm0.042$ & $0.629\pm0.043$   \\
        \midrule
        
        \multicolumn{6}{c}{Coauthor CS}  \\
        \midrule
        $100$ &   $0.883\pm0.019$ & $0.892\pm0.013$ & $0.896\pm0.011$ & $0.899\pm0.008$ & $0.901\pm0.007$   \\
        $200$ &    $0.866\pm0.025$ & $0.885\pm0.015$ & $0.888\pm0.011$ & $0.890\pm0.010$ & $0.890\pm0.010$  \\
        $400$ &   $0.806\pm0.025$ & $0.843\pm0.015$ & $0.848\pm0.014$ & $0.850\pm0.013$ & $0.851\pm0.012$  \\
        $600$ &   $0.807\pm0.030$ & $0.839\pm0.023$ & $0.842\pm0.023$ & $0.844\pm0.022$ & $0.845\pm0.022$  \\
        \bottomrule
    \end{tabular}
    }
\end{table}

\section{The Performance of Variants of the ReLU Activation Function}
\label{sec:variants}

In previous studies, most algorithms have been experimentally analyzed using the ReLU activation function, with limited exploration of its variants. After performing the sensitivity analysis on our proposed algorithm, we further examine the impact of different activation functions by replacing the original ReLU with its variants. In this section, we analyze the test performance of three variants of the ReLU activation function~\cite{nair2010rectified}, such as the Leaky Rectified Linear Unit (Leaky ReLU)~\cite{maas2013rectifier}, Exponential Linear Unit (EReLU)~\cite{DBLP:journals/corr/ClevertUH15}, and the Continuously Differentiable Exponential Linear Unit (CEReLU) function~\cite{barron2017continuously}. All hyperparameter settings are kept consistent with those used for the original ReLU activation function.

\textbf{Leaky ReLU} \cite{maas2013rectifier}: Leaky ReLU is an improved version of the original ReLU activation function, introducing a small linear slope in the negative region ($\alpha x, \alpha \textgreater 0$). Leaky ReLU mitigates the issue of the traditional ReLU having a zero gradient in the negative region. In our experiment, the parameter $\alpha$ was set to its default value of 0.01.

\textbf{EReLU} \cite{DBLP:journals/corr/ClevertUH15}: EReLU produces a smooth output using the exponential function $(\alpha(e^{x} -1), \alpha>0)$ in the negative region, mitigating the zero-gradient issue of ReLU. Additionally, it exhibits a zero-mean property near zero, which facilitates faster network convergence. In our experiment, we set $\alpha = 0.1$.

\textbf{CEReLU} \cite{barron2017continuously}: CEReLU is an enhanced variant of EReLU that introduces an adjustable parameter $\alpha>0$ to ensure the activation function $(\alpha(e^{\frac{x}{\alpha}} - 1), \alpha>0)$ maintains a continuously differentiable first-order derivative in the negative region, making it smoother and more flexible. The parameter $\alpha$ is set to the same value as in the EReLU activation function.

\begin{figure}
\centering
\includegraphics[width=0.8\linewidth]{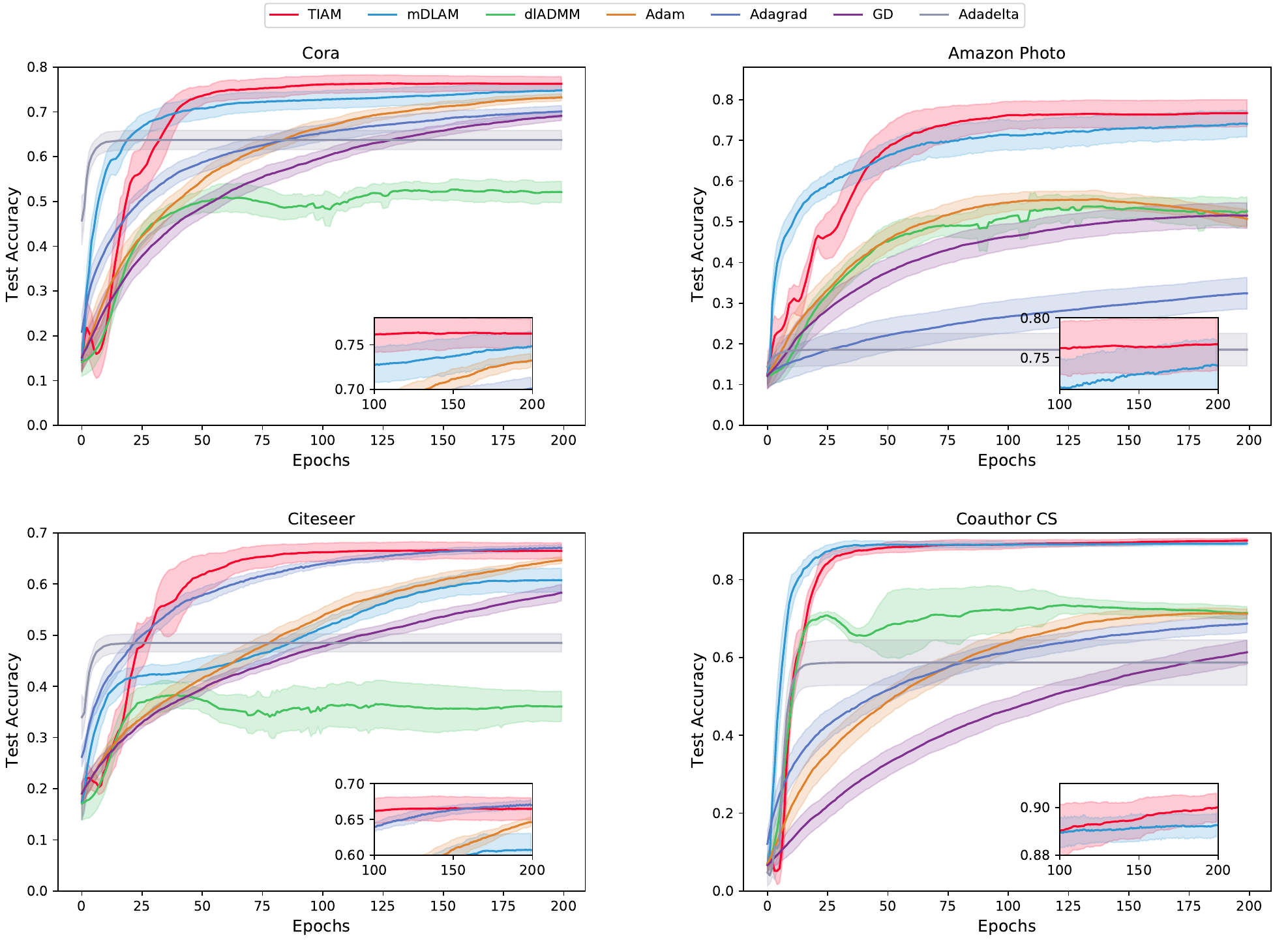}
\caption{The curve illustrates the mean and standard deviation of test accuracy for all methods over 10 runs, utilizing the LeakyReLU activation function.}
\label{fig:leakyrelu_acc}
\end{figure}

\begin{figure}
\centering
\includegraphics[width=0.8\linewidth]{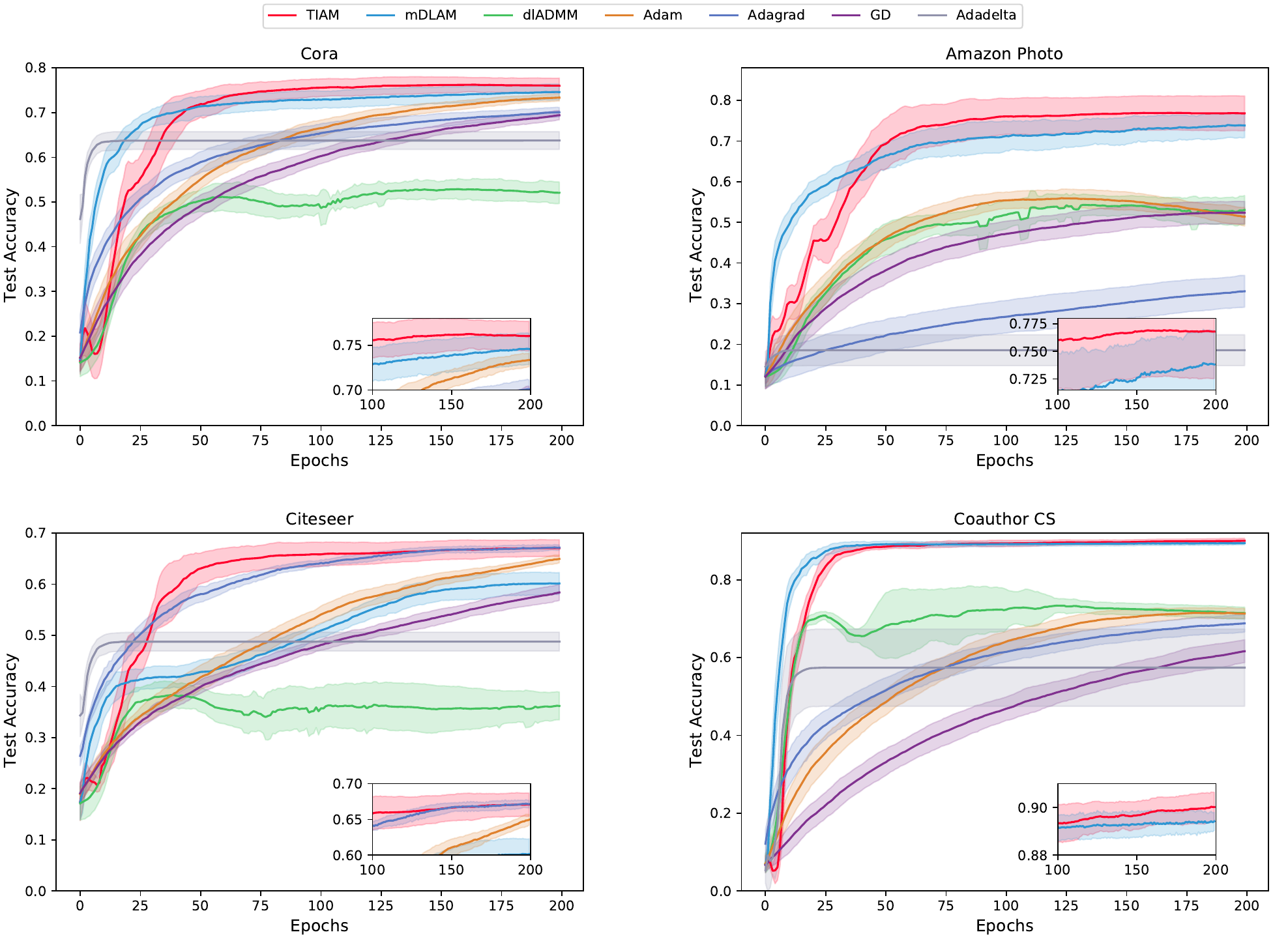}
\caption{The curve illustrates the mean and standard deviation of test accuracy for all methods over 10 runs, utilizing the EReLU activation function. }
\label{fig:erelu_acc}
\end{figure}

\begin{figure}
\centering
\includegraphics[width=0.8\linewidth]{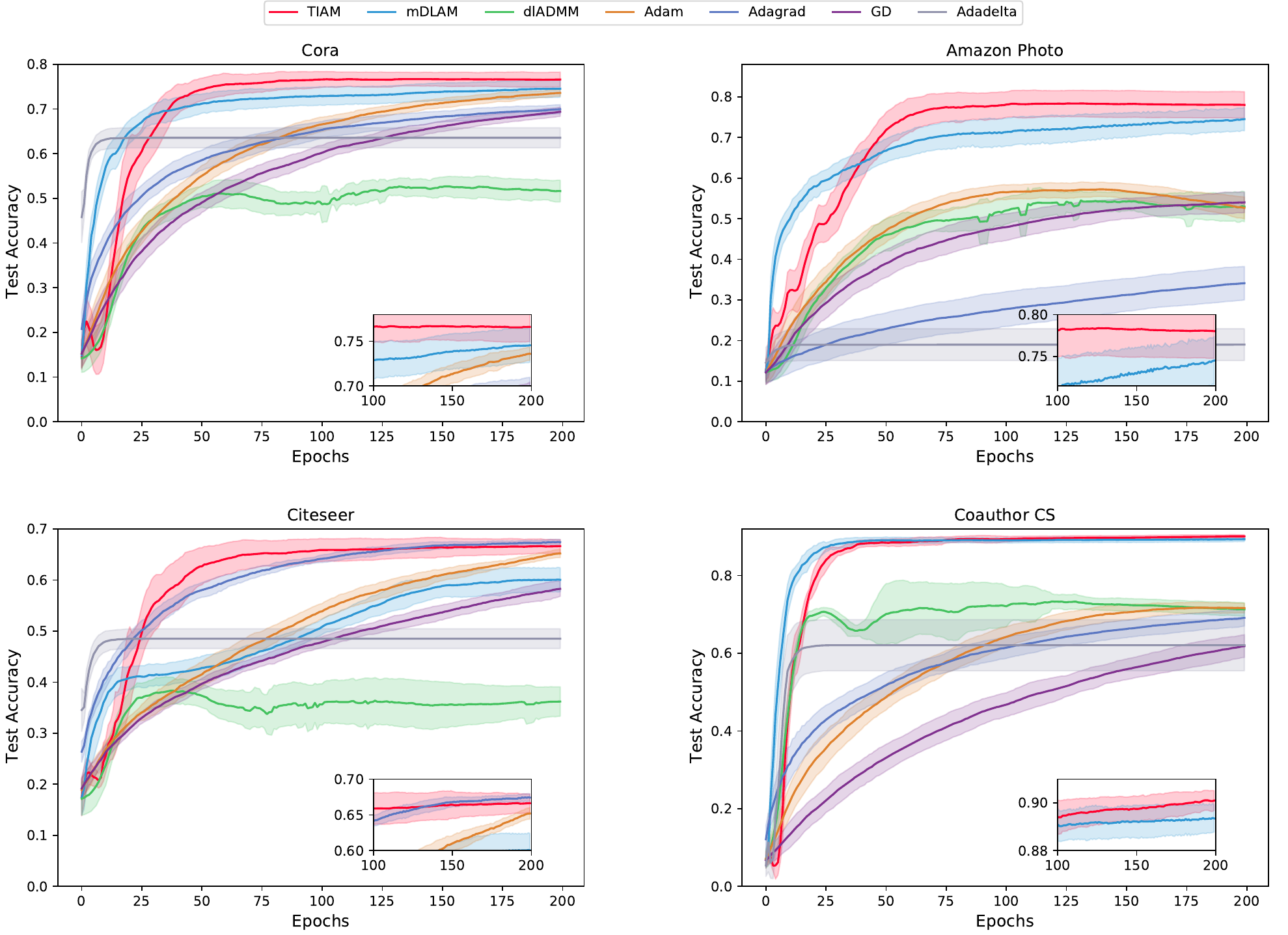}
\caption{The curve illustrates the mean and standard deviation of test accuracy for all methods over 10 runs, utilizing the CEReLU activation function.}
\label{fig:cerelu_acc}
\end{figure}

Figures \ref{fig:leakyrelu_acc}, \ref{fig:erelu_acc}, and \ref{fig:cerelu_acc} illustrate the performance of the TIAM algorithm using different ReLU variants. The results show that the ReLU variants achieve better convergence performance than the standard ReLU activation function on the Cora dataset. However, their test performance on the Citeseer dataset is slightly lower than that of the standard ReLU, yet it still surpasses existing AM-based algorithms. Similarly, altering the activation function has little impact on test performance for the Amazon Photo dataset, as all variants outperform other algorithms within 40 iterations and exhibit minimal fluctuations. For the Coauthor CS dataset, although all activation functions converge slightly more slowly in the early stages, all ReLU variants achieve better test performance later and surpass other algorithms within 75 iterations. Overall, the three ReLU variants perform well across datasets. These findings suggest that the proposed algorithm exhibits better generalization capability and robustness across different activation functions.


\section{Conclusion}\label{conclusion}
\label{sec:Conclusion}

In this work, we proposed a novel AM method with a triple-inertial acceleration framework to solve the MLP optimization problem. Our algorithm combines a specialized approximation approach with a triple-inertial acceleration strategy to enable block-wise accelerated optimization of neural network parameters. Furthermore, we conducted a rigorous convergence analysis and proved that the proposed algorithm achieves a linear convergence rate. Experimental results confirm the better performance and efficiency of our method. In future work, we plan to incorporate stochastic algorithms into the current framework by adopting a stochastic mini-batch approach, which aims further to enhance the efficiency of AM-based neural network optimization algorithms.

\section*{Declarations}
{\bf Funding:} This research is supported by the National Natural Science Foundation of China (NSFC) grants 92473208, 12401415, the Key Program of National Natural Science of China 12331011, the 111 Project (No. D23017),  the Natural Science Foundation of Hunan Province (No. 2025JJ60009), and the Postgraduate Scientific Research Innovation Project of Hunan Province (No. LXBZZ2024114). 

\noindent{\bf Data Availability:} Enquiries about data/code availability should be directed to the authors.

\noindent{\bf Competing interests:} The authors have no competing interests to declare that are relevant to the content of this paper.


\section*{Appendix}
\label{appendix_list}

\subsection*{A.  Proof Preliminary Results}
In this section, we introduce preliminary lemmas for the proposed method. The limiting sub-differential is utilized to establish the convergence of the proposed algorithm in the subsequent analysis. Without loss of generality, we assume that $\partial R$ and $\partial\Omega_l (l=1,\cdots,L)$ are nonempty. The limiting sub-differential of $F(\cdot)$ is defined in the optimization problem \eqref{problem:2}:
$$\partial F(\mathbf{W},\mathbf{z},\mathbf{a},\mathbf{b})=\partial_\mathbf{W}F\times\partial_\mathbf{z}F\times\partial_\mathbf{a}F\times\partial_\mathbf{b}F  ,$$
where $\times$ denotes the Cartesian product.

\begin{lemma}
\label{lemma:pre}
If \eqref{eq:argminW} holds, then there exists $p\in\partial\Omega_l\left(W_l^{k+1}\right)$ such that
$$\nabla_{{W}_l}\phi+\theta_l^{k+1}\Big(W_l^{k+1}-\tilde{W}_l^{k+1}\Big)+p=0,$$
where $\nabla_{{W}_l}\phi\left(\overline{a}_{l-1}^{k+1},\hat{W}_{l}^{k+1},\overline{z}_{l}^{k},\overline{b}_{l}^{k}\right)$ is simplified as $\nabla_{{W}_l}\phi$, and a similar notation applies to the formulas below. 
Likewise, if \eqref{eq:zl_argmin} holds, then there exists $q \in \partial h_l\left(z_{l}^{k+1}\right)$ such that
$$\nabla_{{z}_l}\phi+\rho\left(z_l^{k+1}-\tilde{z}_l^{k+1}\right)+q=0,$$
where $q$ is a subgradient with respect to $z_l^{k+1}$ that satisfies the constraint $h_l(z_l^{k+1})-\epsilon\leqslant a_l^k\leqslant h_l(z_l^{k+1})+\epsilon.$ If \eqref{eq:zL_argmin} holds, then there exists $r\in\partial R(z_L^{k+1};y)$ satisfying

$$\nabla_{{z}_L}\phi+\rho\left(z_L^{k+1}-\tilde{z}_L^{k+1}\right)+r=0.$$

If \eqref{eq:argmina} holds, then there exists $s$ such that
$$\nabla_{{a}_l}\phi+\gamma_l^{k+1}\left(a_l^{k+1}-\tilde{a}_l^{k+1}\right)+s=0,$$
where $s$ is a subgradient with respect to $a_l^{k+1}$ that satisfies the constraint $h_l(z_l^{k+1})-\epsilon \leqslant a_l^{k+1}\leqslant h_l(z_l^{k+1})+\epsilon.$

If \eqref{eq:argminb} holds, with respect to $b_l^{k+1}$, we have
$$\nabla_{{b}_l}\phi+\zeta_l^{k+1}\left(b_l^{k+1}-\tilde{b}_l^{k+1}\right) =0.$$
\end{lemma}

Proof. These results follow directly from the optimality conditions of \eqref{eq:argminW}, \eqref{eq:zL_argmin}, \eqref{eq:argminb}, \eqref{eq:zl_argmin} and \eqref{eq:argmina}, respectively.             \\

Inspired by \cite{bolte2014proximal} and \cite{pock2016inertial}, we make the following assumption:
\begin{assumption}
\label{assu:2}
$\nabla \phi$ is M-Lipchitz continuous on bounded subset of $\mathbb{R} ^{m_1}\times \mathbb{R} ^{m_2}\times \mathbb{R} ^{m_3}\times \mathbb{R} ^{m_4}$. In other words, for each bounded subset $B_1~\times~B_2~\times~B_3\times~B_4\subset~\mathbb{R}^{m_1}\times~\mathbb{R}^{m_2}\times~\mathbb{R}^{m_3}\times~\mathbb{R}^{m_4}$, there exists $M > 0$ such that for all $(a^k,W^k,z^k,b^k) \in B_1\times B_2\times B_3\times B_4, \left(i=1,2,\cdots,N\right)$,
\begin{align*}
\left.\left\|\begin{pmatrix}\nabla_a\phi(a^1,W^1,z^1,b^1)\\\nabla_W\phi(a^1,W^1,z^1,b^1)\\\nabla_z\phi(a^1,W^1,z^1,b^1)\\\nabla_{b}\phi(a^1,W^1,z^1,b^1)\end{pmatrix}\right.-\begin{pmatrix}\nabla_{a}\phi(a^2,W^2,z^2,b^2)\\\nabla_{W}\phi(a^2,W^2,z^2,b^2)\\\nabla_{z}\phi(a^2,W^2,z^2,b^2)\\\nabla_{b}\phi(a^2,W^2,z^2,b^2)\end{pmatrix}\right\|\leq M\left\|\begin{pmatrix}a^1\\W^1\\z^1\\b^1\end{pmatrix}-\begin{pmatrix}a^2\\W^2\\z^2\\b^2\\\end{pmatrix}\right\|.
\end{align*}
\end{assumption}

\subsection*{B.  Proof of Lemma}
\label{lemma:final_result}

\noindent\textbf{Proof of Lemma} \ref{lemma:F} \\
\textbf{Proof. } In the parameter update step for $W$ in Algorithm \ref{alg:Inertial}, if $F\left(\mathbf{W}_{ l}^{k+1},\mathbf{z}_{l-1}^{k+1},\mathbf{a}_{ l-1}^{k+1},\mathbf{b}_{ l-1}^{k+1}\right)<$ $F\left(\mathbf{W}_{ l-1}^{k+1},\mathbf{z}_{ l-1}^{k+1},\mathbf{a}_{ l-1}^{k+1},\mathbf{b}_{l-1}^{k+1}\right)$, then it is evident that there exists $\beta_l^{k+1}>0$ such that (\ref{eq:Conver_ineq1}) holds. Otherwise, if the objective function $F(\cdot)$ increases after updating $W$, since $\Omega_{l}(W_l)$ is convex with respect to $W_l$, it follows from the definition of the regular subgradient and \eqref{eq:leqwphi} that
\begin{align}
\label{eq:w_phi}
&\Omega_{l}\left(W_{l}^{k}\right)\geqslant\Omega_{l}\left(W_{l}^{k+1}\right)+p^{T}\left(W_{l}^{k}-W_{l}^{k+1}\right),   
\end{align}
where $p \in \partial\Omega_l\left(W_l^{k+1}\right)$ is defined in Lemma \ref{lemma:pre}. For the parameter $W$, if the objective function $F(\cdot)$ increases after updating $W$, based on line 6 of Algorithm \ref{alg:Inertial}, all accelerated variables revert to their non-accelerated states for a repeated update. Therefore, we have
\begin{align*}
&F\left(\mathbf{W}_{l-1}^{k+1},\mathbf{z}_{l-1}^{k+1},\mathbf{a}_{l-1}^{k+1},\mathbf{b}_{l-1}^{k+1}\right)-F\left(\mathbf{W}_{l}^{k+1},\mathbf{z}_{l-1}^{k+1},\mathbf{a}_{l-1}^{k+1},\mathbf{b}_{l-1}^{k+1}\right)  \\
&=\phi\left(a_{l-1}^{k+1},W_l^k,z_l^k,b_l^k\right)+\Omega_l\left(W_l^k\right)-\phi\left(a_{l-1}^{k+1},W_l^{k+1},z_l^k,b_l^k\right) - \Omega_l\left(W_l^{k+1}\right) 
 \gray{ \text{(Definition of $F(\cdot)$)} }\\
&\geqslant\Omega_l\left(W_l^k\right)-\Omega_l\left(W_l^{k+1}\right)-\left(\nabla_{{W}_l} \phi\left(\overline{a}_{l-1}^{k+1},\hat{W}_l^{k+1},\overline{z}_l^k,\overline{b}_l^k\right) \right)^{T}\left(W_l^{k+1}-\tilde{W}_l^{k+1}\right) -\frac{\theta_l^{k+1}}{2}\|W_l^{k+1}-\tilde{W}_l^{k+1}\|^2 \\
& - \phi\left(\overline{a}_{l-1}^{k+1},\tilde{W}_l^{k+1},\overline{z}_l^k,\overline{b}_l^k\right)+\phi\left(a_{l-1}^{k+1},W_l^k,z_l^k,b_l^k\right)    .       \gray{ \text{ (See \eqref{eq:w_phi}, \eqref{eq:leqwphi} and Algorithm \ref{alg:Inertial}) } }   \\
&\geqslant p^T\left(W_l^k-W_l^{k+1}\right)-\left(\nabla_{{W}_l} \phi\left(\overline{a}_{l-1}^{k+1},\hat{W}_l^{k+1},\overline{z}_l^k,\overline{b}_l^k\right) \right)^T\left(W_l^{k+1}-\tilde{W}_l^{k+1}\right)-\frac{\theta_l^{k+1}}{2}\|W_l^{k+1}-\tilde{W}_l^{k+1}\|^2   \\
 & =-\left(\nabla_{{W}_{l}} \phi\left(\overline{a}_{l-1}^{k+1},\hat{W}_l^{k+1},\overline{z}_l^k,\overline{b}_l^k\right) + \theta_{l}^{k+1}\left( W_{l}^{k+1} - \tilde{W}_{l}^{k+1} \right) \right)^{T} \left(W_{l}^{k}-W_{l}^{k+1}\right) \\
 & -\left(\nabla_{{W}_l} \phi\left(\overline{a}_{l-1}^{k+1},\hat{W}_l^{k+1},\overline{z}_l^k,\overline{b}_l^k\right) 
 \right)^T\left(W_l^{k+1}-W_l^k\right)-\frac{\theta_l^{k+1}}{2}\|W_l^{k+1}-\tilde{W}_l^{k+1}\|^2   \gray{ \text{(  Lemma~\ref{lemma:pre}, Algorithm \ref{alg:Inertial}) }} \\
 & = \theta_l^{k+1}\left(W_l^{k+1}-\tilde{W}_l^{k+1}\right)^T\left( {W}_l^{k+1}-W_l^k\right) - \frac{\theta_l^{k+1}}{2}\|W_l^{k+1}-\tilde{W}_l^{k+1}\|^2 \\
 & =\frac{\theta_l^{k+1}}{2}\|W_l^{k+1}-W_l^k\|^2   \gray{ 
 \left(\tilde{W}_l^{k+1}=W_l^k\right) }.
\end{align*}
Let $\theta_l^{k+1} = \beta_l^{k+1}$, then \eqref{eq:Conver_ineq1} still holds.   \\


\noindent\textbf{Proof of Lemma} \ref{lemma:decrease} \\
\textbf{Proof. } Here, we add \eqref{eq:Conver_ineq1}, \eqref{eq:Conver_ineq2}, \eqref{eq:Conver_ineq3} and \eqref{eq:Conver_ineq4} from $l = 1 ~ to ~ L$ to obtain

\begin{align*}
&F\left(\mathbf{W}^{k},\mathbf{z}^{k},\mathbf{a}^{k},\mathbf{b}^{k}\right) -F\left(\mathbf{W}^{k+1},\mathbf{z}^{k+1},\mathbf{a}^{k+1},\mathbf{b}^{k+1}\right)  \\
&\geqslant \sum_{l=1}^{L}\left(\frac{\beta_{l}^{k+1}}{2}\|W_{l}^{k+1}-W_{l}^{k}\|^{2}+\frac{\delta_{l}^{k+1}}{2}\|z_{l}^{k+1}-z_{l}^{k}\|^{2} +\frac{\zeta_{l}^{k+1}}{2}\|b_{l}^{k+1}-b_{l}^{k}\|^{2} \right)+\sum_{l=1}^{L-1}\frac{\gamma_{l}^{k+1}}{2}\|a_{l}^{k+1}-a_{l}^{k}\|^{2}      .
\end{align*}
Let $C_{5}=\operatorname*{min}\left({\frac{\beta_{l}^{k+1}}{2}},{\frac{ \delta_{l}^{k+1}}{2}},{\frac{\zeta_{l}^{k+1}}{2}},{\frac{\gamma_{l}^{k+1}}{2}} \right)>0$, we have
\begin{align}
\label{eq:FK-FK+1}
&F\Big(\mathbf{W}^{k},\mathbf{z}^{k},\mathbf{a}^{k},\mathbf{b}^{k}\Big)-F\Big(\mathbf{W}^{k+1},\mathbf{z}^{k+1},\mathbf{a}^{k+1},\mathbf{b}^{k+1}\Big)                   \nonumber \\
&\geqslant C_5\left(\sum_{l=1}^L\left(\|W_l^{k+1}-W_l^k\|^2+\|z_l^{k+1}-z_l^k\|^2 + \|b_l^{k+1}-b_l^k\|^2 \right)+\sum_{l=1}^{L-1}\|a_l^{k+1}-a_l^k\|^2\right)   \nonumber \\
&=C_5\left(\|\mathbf{W}^{k+1}-\mathbf{W}^k\|^2+\|\mathbf{z}^{k+1}-\mathbf{z}^k\|^2+\|\mathbf{b}^{k+1}-\mathbf{b}^k\|^2+\|\mathbf{a}^{k+1}-\mathbf{a}^k\|^2\right) \geqslant 0.          
\end{align}
Since all variables (i.e. $\mathbf{W}^{k},\mathbf{z}^{k},\mathbf{a}^{k},\mathbf{b}^{k}$) are bounded (Lemma \ref{lemma:bounder}), it follows that $F\left(\mathbf{W}^{k},\mathbf{z}^{k},\mathbf{a}^{k},\mathbf{b}^{k}\right)$ is convergent. \\


\noindent \textbf{Proof of Lemma} \ref{lemma:bounder} \\
\textbf{Proof. } Firstly, we sum \eqref{eq:Conver_ineq1}, \eqref{eq:Conver_ineq2}, \eqref{eq:Conver_ineq3}, and \eqref{eq:Conver_ineq4} over $l=1$ to $L$ and $k=0$ to $K$ to obtain

\begin{align}
\label{eq:boundproof}
&F(\mathbf{W}^{0},\mathbf{z}^{0},\mathbf{a}^{0},\mathbf{b}^{0})  - F(\mathbf{W}^{k},\mathbf{z}^{k},\mathbf{a}^{k},\mathbf{b}^{k}) \geqslant   \nonumber \\
&\sum_{k=0}^{K}\left(  \sum_{l=1}^{L}\left(\frac{\beta_{l}^{k+1}}{2}\|W_{l}^{k+1}-W_{l}^{k}\|^{2}+\frac{\delta_{l}^{k+1}}{2}\|z_{l}^{k+1}-z_{l}^{k}\|^{2} +\frac{\zeta_{l}^{k+1}}{2}\|b_{l}^{k+1}-b_{l}^{k}\|^{2} \right)+\sum_{l=1}^{L-1}\frac{\gamma_{l}^{k+1}}{2}\|a_{l}^{k+1}-a_{l}^{k}\|^{2}  \right)  ,
\end{align}
then we have $F(\mathbf{W}^{0},\mathbf{z}^{0},\mathbf{a}^{0},\mathbf{b}^{0})  \geqslant F(\mathbf{W}^{k},\mathbf{z}^{k},\mathbf{a}^{k},\mathbf{b}^{k})$, which establishes the upper bound of $F(\cdot)$. Let $k\to\infty$ in \eqref{eq:boundproof}, and noting that $F>0$ serves as a lower bound, we obtain

\begin{align*}
    &\sum_{k=0}^{K}\left(  \sum_{l=1}^{L}\left(\frac{\beta_{l}^{k+1}}{2}\|W_{l}^{k+1}-W_{l}^{k}\|^{2}+\frac{\delta_{l}^{k+1}}{2}\|z_{l}^{k+1}-z_{l}^{k}\|^{2} +\frac{\zeta_{l}^{k+1}}{2}\|b_{l}^{k+1}-b_{l}^{k}\|^{2} \right)+\sum_{l=1}^{L-1}\frac{\gamma_{l}^{k+1}}{2}\|a_{l}^{k+1}-a_{l}^{k}\|^{2}  \right)   \\
    &\textless \infty .
\end{align*}
Since the sum of this infinite series is finite, each term converges to 0. This means that $\operatorname*{lim}_{k\to\infty}\|W_{l}^{k+1}-W_{l}^{k}\|=0,\operatorname*{lim}_{k\to\infty}\|z_{l}^{k+1}-z_{l}^{k}\|=0,\operatorname*{lim}_{k\to\infty}\|b_{l}^{k+1}-b_{l}^{k}\|=0 \mathrm{~and~} \operatorname*{lim}_{k\to\infty}\|a_{l}^{k+1}-a_{l}^{k}\|=0$, which proves that $\lim_{k\to\infty}\|\mathbf{W}^{k+1}-\mathbf{W}^{k}\|=0, \lim_{k\to\infty}\|\mathbf{z}^{k+1}-\mathbf{z}^{k}\|=0, \lim_{k\to\infty}\|\mathbf{b}^{k+1}-\mathbf{b}^{k}\|=0, \mathrm{~and~} \lim_{k\to\infty}\|\mathbf{a}^{k+1}-\mathbf{a}^{k}\|=0$.

Secondly, since $F\left(\mathbf{W}^k,\mathbf{z}^k,\mathbf{a}^k,\mathbf{b}^k\right)$ is bounded, it follows from the definition of coercivity and Assumption \ref{assu:1} that $\left(\mathbf{W}^k,\mathbf{z}^k,\mathbf{a}^k,\mathbf{b}^k\right)$ is also bounded.   \\


\noindent \textbf{Proof of Lemma} \ref{lemma:subgradient} \\
\textbf{Proof. } In this part, we analyze the subgradient bound in our method. Taking the parameter $\mathbf{W}$ as an example, we define

$$\partial_{\mathbf{W}^{k+1}}F=\left\{\partial_{W_{1}^{k+1}}F\right\}\times\left\{\partial_{W_{2}^{k+1}}F\right\}\times\cdots\times\left\{\partial_{W_{L}^{k+1}}F\right\},$$
where $\times$ denotes the Cartesian product.

In Algorithm \ref{algo:block}, if ${F}(\mathbf{W}_{l}^{k+1},\mathbf{z}_{l-1}^{k+1},\mathbf{a}_{l-1}^{k+1},\mathbf{b}_{l-1}^{k+1}) \geq {F}(\mathbf{W}_{l-1}^{k+1},\mathbf{z}_{l-1}^{k+1},\mathbf{a}_{l-1}^{k+1},\mathbf{b}_{l-1}^{k+1})$, then according to the definition of $F(\cdot)$ in the optimization problem \eqref{problem:2}, we have
\begin{align*}
&\partial_{W_{l}^{k+1}}F=\partial\Omega_{l}\Big(W_{l}^{k+1}\Big)+\nabla_{W_{l}}\phi\Big(a_{l-1}^{k+1},W_{l}^{k+1},z_{l}^{k+1},b_{l}^{k+1}\Big) \\
&=\nabla_{W_l}\phi\left(a_{l-1}^{k+1},W_l^{k+1},z_l^{k+1},b_{l}^{k+1}\right)-\nabla_{{W}_l}\phi\left(\overline{a}_{l-1}^{k+1},\hat{W}_l^{k+1},\overline{z}_l^k,\overline{b}_{l}^{k}\right) -\theta_l^{k+1}\left(W_l^{k+1}- \tilde{W}_l^{k+1}\right) \\ &+\partial\Omega_l\left(W_l^{k+1}\right)+\nabla_{{W}_l}\phi\left(\overline{a}_{l-1}^{k+1},\hat{W}_l^{k+1},\overline{z}_l^k,\overline{b}_{l}^{k}\right)+\theta_l^{k+1}\left(W_l^{k+1}-\tilde{W}_l^{k+1}\right)   .
\end{align*}
By applying the triangle inequality and Assumption \ref{assu:2}, there exists a constant $M$ such that
\begin{align*}
&\| \nabla_{W_l}\phi\left(a_{l-1}^{k+1},W_l^{k+1},z_l^{k+1},b_{l}^{k+1}\right)-\nabla_{{W}_l}\phi\left(\overline{a}_{l-1}^{k+1},\hat{W}_l^{k+1},\overline{z}_l^k,\overline{b}_{l}^{k}\right) -\theta_l^{k+1}\left(W_l^{k+1}- \tilde{W}_l^{k+1}\right) \|   \\
&\leq  \| \nabla_{W_l}\phi\left(a_{l-1}^{k+1},W_l^{k+1},z_l^{k+1},b_{l}^{k+1}\right)-\nabla_{{W}_l}\phi\left(\overline{a}_{l-1}^{k+1},\hat{W}_l^{k+1},\overline{z}_l^k,\overline{b}_{l}^{k}\right)\| + \theta_l^{k+1} \|\left(W_l^{k+1}- \tilde{W}_l^{k+1}\right) \| .  \\
&\leq   M\left(   \|a_{l-1}^{k+1} - \overline{a}_{l-1}^{k+1} \| + \|W_{l}^{k+1} - \hat{W}_{l}^{k+1} \| + \|z_{l}^{k+1} - \overline{z}_{l}^{k} \| + \|b_{l}^{k+1} - \overline{b}_{l}^{k} \|   \right) + \theta_l^{k+1} \| W_l^{k+1}- \tilde{W}_l^{k+1} \|    \\
&\leq   M\Big(   \|a_{l-1}^{k+1} - \left(  a_{l-1}^{k+1}+p_{3}\left(a_{l-1}^{k+1}-a_{l-1}^{k}\right)  \right) \| + \|W_{l}^{k+1} - \left(   W_{l}^{k}+p_{2}\left(W_{l}^{k}-W_{l}^{k-1}\right)  \right) \|   \\
&+ \|z_{l}^{k+1} - \left(  z_{l}^{k}+p_{3}\left(z_{l}^{k}-z_{l}^{k-1}\right)  \right) \| + \|b_{l}^{k+1} - \left(  b_{l}^{k}+p_{3}\left(b_{l}^{k}-b_{l}^{k-1}\right)  \right) \|   \Big)  \\
&+ \theta_l^{k+1} \| W_l^{k+1}- \left(  W_{l}^{k}+p_{1}\left(W_{l}^{k}-W_{l}^{k-1}\right)  \right) \|  . \quad   \gray{  \text{(Algorithm~\ref{alg:Inertial})}  } \\
&\leq   M\Big(  \| a_{l-1}^{k+1}-a_{l-1}^{k} \|   + \| W_{l}^{k+1} -  W_{l}^{k}\| + \| W_{l}^{k} -  W_{l}^{k-1}\| + \| z_{l}^{k+1} -  z_{l}^{k}\| + \| z_{l}^{k} -  z_{l}^{k-1}\|   \\
&  +   \| b_{l}^{k+1} -  b_{l}^{k}\| + \| b_{l}^{k} -  b_{l}^{k-1}\|   \Big)  + \theta_l^{k+1} \left( \| W_{l}^{k+1} -  W_{l}^{k}\| + \| W_{l}^{k} -  W_{l}^{k-1}\| \right)   ,
\end{align*}
where the last inequality follows from the conditions $0\leq p_{1}<1 , 0\leq p_{2}<1 \text{, and } 0\leq p_{3}<1$. Returning to the original format of $\partial_{\mathbf{W}^{k+1}}F$, and based on the optimality condition in \eqref{eq:argminW}, we obtain
\begin{align*}
0 \in \partial\Omega_l\left(W_l^{k+1}\right)+\nabla_{{W}_l}\phi\left(\overline{a}_{l-1}^{k+1},\hat{W}_l^{k+1},\overline{z}_l^k,\overline{b}_{l}^{k}\right)+\theta_l^{k+1}\left(W_l^{k+1}-\tilde{W}_l^{k+1}\right)   .
\end{align*}
Therefore, there exists $g_{1,l}^{k+1} \in \partial_{W_{l}^{k+1}}F$ such that
\begin{align*}
\|g_{1,l}^{k+1}\| & \leqslant \left( M + \theta_l^{k+1} \right)\|W_l^{k+1}-W_l^k\| +\left( M + \theta_l^{k+1} \right)\|W_l^k-W_l^{k-1}\| + M \|b_{l}^{k+1}-b_{l}^{k}\| \\
&+ M \|b_{l}^{k}-b_{l}^{k-1}\| + M \|z_l^{k+1}-z_l^k\|  + M \|z_l^{k}-z_l^{k-1}\| + M \|a_{l-1}^{k+1}-a_{l-1}^{k}\| .
\end{align*}
So there exists $g_1^{k+1}=g_{1,1}^{k+1}\times g_{1,2}^{k+1}\times\cdots\times g_{1,L}^{k+1}\in\partial_{\mathbf{W}^{k+1}}F$, and the coefficient of the corresponding subgradient is $C_1=\max\left( M , M +\theta_{1}^{k+1}, M +\theta_{2}^{k+1},\cdots,  M +\theta_{L}^{k+1}\right)$ such that

\begin{align}
\label{eq:subgradientW1}
\|g_1^{k+1}\| & \leqslant C_1\Big(\|\mathbf{W}^{k+1}-\mathbf{W}^k\|+\|\mathbf{z}^{k+1}-\mathbf{z}^k\|+ \|\mathbf{a}^{k+1}-\mathbf{a}^k\| +\|\mathbf{b}^{k+1}-\mathbf{b}^k\|+\|\mathbf{W}^k-\mathbf{W}^{k-1}\| +\|\mathbf{b}^k-\mathbf{b}^{k-1}\| + \|\mathbf{z}^k-\mathbf{z}^{k-1}\| \Big)   .
\end{align}

Otherwise, if the objective function $F(\cdot)$ increases after updating the parameter $\mathbf{W}$, applying the triangle inequality, Assumption \ref{assu:2}, and line 6 of Algorithm \ref{alg:Inertial}, there exists $M$ such that
\begin{align*}
&\| \nabla_{W_l}\phi\left(a_{l-1}^{k+1},W_l^{k+1},z_l^{k+1},b_{l}^{k+1}\right)-\nabla_{{W}_l}\phi\left(\overline{a}_{l-1}^{k+1},\hat{W}_l^{k+1},\overline{z}_l^k,\overline{b}_{l}^{k}\right) -\theta_l^{k+1}\left(W_l^{k+1}- \tilde{W}_l^{k+1}\right) \|   \\
&\leq  \| \nabla_{W_l}\phi\left(a_{l-1}^{k+1},W_l^{k+1},z_l^{k+1},b_{l}^{k+1}\right)-\nabla_{{W}_l}\phi\left(\overline{a}_{l-1}^{k+1},\hat{W}_l^{k+1},\overline{z}_l^k,\overline{b}_{l}^{k}\right)\| + \theta_l^{k+1} \|\left(W_l^{k+1}- \tilde{W}_l^{k+1}\right) \| .  \\
&\leq   M\left(   \|a_{l-1}^{k+1} - \overline{a}_{l-1}^{k+1} \| + \|W_{l}^{k+1} - \hat{W}_{l}^{k+1} \| + \|z_{l}^{k+1} - \overline{z}_{l}^{k} \| + \|b_{l}^{k+1} - \overline{b}_{l}^{k} \|   \right) + \theta_l^{k+1} \| W_l^{k+1}- \tilde{W}_l^{k+1} \|    \\
&\leq   M\Big(   \|W_{l}^{k+1} -   W_{l}^{k} \| + \|z_{l}^{k+1} -  z_{l}^{k}  \|  + \|b_{l}^{k+1} -  b_{l}^{k}  \|   \Big)  + \theta_l^{k+1} \| W_l^{k+1}-  W_{l}^{k}  \|  .      \quad   \gray{   \text{(Algorithm~\ref{alg:Inertial} )}   } 
\end{align*}
Under this condition, the optimality condition of \eqref{eq:argminW} has
$$0 \in \partial\Omega_l\left(W_l^{k+1}\right)+\nabla_{{W}_l}\phi\left(a_{l-1}^{k+1},\hat{W}_l^{k+1},z_l^k,b_{l}^{k}\right)+\theta_l^{k+1}\left(W_l^{k+1}-\tilde{W}_l^{k+1}\right).$$
Similarly, we obtain
\begin{align}
\label{eq:subgradientW2}
\|g_{1,l}^{k+1}\| & \leqslant \left( M + \theta_l^{k+1} \right)\|W_l^{k+1}-W_l^k\|  + M \|z_l^{k+1}-z_l^k\|  + M \|b_{l}^{k+1}-b_{l}^{k}\|  .
\end{align}

By combining \eqref{eq:subgradientW1} and \eqref{eq:subgradientW2}, we obtain $C_{1}= \max\left( M , M +\theta_{1}^{k+1}, M +\theta_{2}^{k+1},\cdots,  M +\theta_{L}^{k+1}\right)$ and $ g_{1}^{k+1}=g_{1,1}^{k+1}\times g_{1,2}^{k+1}\times\cdots\times g_{1,L}^{k+1}\in\partial_{\mathbf{W}^{k+1}}F$ such that
\begin{align}
\label{eq:proofsubgradientw_l}
\|g_1^{k+1}\| & \leqslant C_1\Big(\|\mathbf{W}^{k+1}-\mathbf{W}^k\|+\|\mathbf{z}^{k+1}-\mathbf{z}^k\|+ \|\mathbf{a}^{k+1}-\mathbf{a}^k\| +\|\mathbf{b}^{k+1}-\mathbf{b}^k\|+\|\mathbf{W}^k-\mathbf{W}^{k-1}\| + \|\mathbf{b}^k-\mathbf{b}^{k-1}\| +\|\mathbf{z}^k-\mathbf{z}^{k-1}\| \Big)   .
\end{align}

\subsection*{C.  Proof of Theorem}
\label{converge:final_result}

\noindent \textbf{Proof of Theorem} \ref{theory:conver} \\
\textbf{Proof. } In this part, we now begin to explore the convergence of our method. Taking the parameter $\mathbf{W}$ as an example, by Lemma \ref{lemma:bounder}, we have $\operatorname*{lim}_{k\to\infty}\|\mathbf{W}^{k+1}-\mathbf{W}^{k}\|=0$, and there exists a subsequence $\mathbf{W}^s$ such that $\mathbf{W}^s\to \mathbf{W}^*$ , where $\mathbf{W}^*$ is a limit point. By Lemma \ref{lemma:subgradient}, there exists $g_{1}^{s}\in\partial_{\mathbf{W}^{s}}F$ such that $\|\mathrm{g}_1^s\|\to0\mathrm{~as~}s\to\infty$. From the definition of the limiting sub-differential, there exists $0 \in \partial_{\mathbf{W}^{s}}F$, which implies that $\mathbf{W}^{*}$ is a stationary point of the objective function $F(\cdot)$ in the optimization problem \eqref{problem:2}.    \\

\noindent \textbf{Proof of Theorem} \ref{theory:rate} \\
\textbf{Proof. } First, let us review the definition of the KL property:

A function $f(x)$ has the KL Property at $\overline{x}\in dom$ $\partial f=\{x\in\mathbb{R}:\partial f(x)\neq\varnothing\}$ if there exists $\eta\in(0,+\infty]$, a neighborhood $X$ of $\bar{x}$ and a function $\psi\in\Psi_\eta$, such that for all

$$x\in X\cap\{x\in\mathbb R:f(\overline{x})<f(x)<f(\overline{x})+\eta\},$$
the following inequality holds
$$\psi'(f(x)-f(\overline{x})) ~ dist(0,\partial f(x))\geqslant1,$$
where $\Psi_\eta$ stands for a class of function $\psi:[0,\eta]\to\mathbb{R}^+$ satisfying: \\
\begin{itemize}
    \item[(1)] $\phi$ is concave and $\psi^\prime(x)$ continuous on $(0,\eta)$
    \item[(2)] $\psi$ is continuous at $0,\psi(0)=0;$
    \item[(3)] $\psi^\prime(x)>0,\forall x\in(0,\eta).$ 
\end{itemize}

Lemma \ref{lemma:kl} ~\cite{xu2013block} shows that a locally strongly convex function satisfies the KL Property:
\begin{lemma}
\label{lemma:kl}
A locally strongly convex function $f(x)$ with a constant $\mu$ satisfies the KL Property at any $x\in\mathbb{D}$ with $\psi(x)=\frac2\mu\sqrt{x}$ $\operatorname{and}X=\mathbb{D}\cap\{y:f(y)\geqslant f(x)\}.$
\end{lemma}

Second, we proceed to prove the relevant properties of the other parameters.

(1) For $b_l^{k+1}$, we have 
\begin{align*}
 &\partial_{b_{l}^{k+1}}F  = \nabla_{b_{l}}\phi(a_{l-1}^{k+1},W_{l}^{k+1},z_{l}^{k+1},b_{l}^{k+1}) \\
 & =\nabla_{b_{l}}\phi(a_{l-1}^{k+1},W_{l}^{k+1},z_{l}^{k+1},b_{l}^{k+1}) - \rho(b_{l}^{k+1}-\tilde{b}_{l}^{k+1})-\nabla_{{b}_{l}}\phi(\overline{a}_{l-1}^{k+1},\overline{W}_{l}^{k+1},\overline{z}_{l}^{k+1},\hat{b}_{l}^{k+1}) \\
 &+\nabla_{{b}_{l}}\phi(\overline{a}_{l-1}^{k+1},\overline{W}_{l}^{k+1},\overline{z}_{l}^{k+1},\hat{b}_{l}^{k+1})+\rho(b_{l}^{k+1}-\tilde{b}_{l}^{k+1})  .
\end{align*}
Based on Assumption \ref{assu:2}, Algorithm \ref{alg:Inertial}, and the triangle inequality, we obtain
\begin{align*}
&\| \nabla_{b_{l}}\phi(a_{l-1}^{k+1},W_{l}^{k+1},z_{l}^{k+1},b_{l}^{k+1}) -\nabla_{{b}_{l}}\phi(\overline{a}_{l-1}^{k+1},\overline{W}_{l}^{k+1},\overline{z}_{l}^{k+1},\hat{b}_{l}^{k+1}) -\rho(b_{l}^{k+1}-\tilde{b}_{l}^{k+1}) \|       \\
&\leq   M\left(   \|a_{l-1}^{k+1} - \overline{a}_{l-1}^{k+1} \| + \|W_{l}^{k+1} - \overline{W}_{l}^{k+1} \| + \|z_{l}^{k+1} - \overline{z}_{l}^{k+1} \| + \|b_{l}^{k+1} - \hat{b}_{l}^{k+1} \|   \right) + \rho \| b_l^{k+1}- \tilde{b}_l^{k+1} \|    \\
& \leq   M\Big(  \|a_{l-1}^{k+1} -  \left(  a_{l-1}^{k+1}+p_{3}\left(a_{l-1}^{k+1}-a_{l-1}^{k}\right) \right)  \| + \|W_{l}^{k+1} - \left  (W_{l}^{k+1}+p_{3}\left(W_{l}^{k+1}-W_{l}^{k}\right)  \right)   \|     \\
& + \|z_{l}^{k+1} - \left(   z_{l}^{k+1}+p_{3}\left(z_{l}^{k+1}-z_{l}^{k}\right)  \right) \| + \|b_{l}^{k+1} - \left(  b_{l}^{k}+p_{2}\left(b_{l}^{k}-b_{l}^{k-1}\right)  \right) \|  \Big)      \\
&+  \rho \| b_l^{k+1}- \left(  b_{l}^{k}+p_{1}\left(b_{l}^{k}-b_{l}^{k-1}\right)  \right) \|      \\
& \leq M \Big(    \|  a_{l-1}^{k+1}-a_{l-1}^{k}  \|  +  \|  W_{l}^{k+1}-W_{l}^{k}  \| + \|  z_{l}^{k+1}-z_{l}^{k}  \|  + \| b_{l}^{k+1} - b_{l}^{k}   \| + \|  b_{l}^{k} - b_{l}^{k-1}   \|   \Big) + \rho \| b_{l}^{k+1} - b_{l}^{k} \|    \\ 
& + \rho  \|  b_{l}^{k} - b_{l}^{k-1}   \|    . 
\end{align*} 

Since the optimality condition of \eqref{eq:argminb} leads to
$$ 0 \in   \nabla_{{b}_{l}}\phi(\overline{a}_{l-1}^{k+1},\overline{W}_{l}^{k+1},\overline{z}_{l}^{k+1},\hat{b}_{l}^{k+1})+\rho(b_{l}^{k+1}-\tilde{b}_{l}^{k+1})   ,$$
there exists $g_{6,l}^{k+1}\in\partial_{b_l^{k+1}}F$ such that 
$$
\|g_{6,l}^{k+1}\| \leq \left( M+\rho \right) \|b_l^{k+1} -  b_l^{k}  \| + \left( M+\rho \right) \| b_l^{k} -  b_l^{k-1}  \| + M  \|  a_{l-1}^{k+1}-a_{l-1}^{k}  \| + M  \|  W_{l}^{k+1}-W_{l}^{k}  \| + M \|  z_{l}^{k+1}-z_{l}^{k}  \|    .
$$
Therefore, there exists $g_6^{k+1}=g_{6,1}^{k+1}\times g_{6,2}^{k+1}\times\cdots\times g_{6,L}^{k+1}\in\partial_{\mathbf{b}^{k+1}}F$ and the coefficient of correspond subgradient is $C_6 = \max\left( M , M + \rho \right)$ such that
\begin{align}
\label{eq:subgradientb1}
\|g_6^{k+1}\|\leqslant C_6\Big(\|\mathbf{b}^{k+1}-\mathbf{b}^k\|+ |\mathbf{b}^{k}-\mathbf{b}^{k-1}\| + \|\mathbf{W}^{k+1}-\mathbf{W}^k\|  + \|\mathbf{z}^{k+1}-\mathbf{z}^k\| + \|\mathbf{a}^{k+1}-\mathbf{a}^k\|\Big)    .
\end{align}

Otherwise, if the objective function $F(\cdot)$ increases after updating parameter $b$, following $ \tilde{b}_{l}^{k+1} = b_{l}^{k}, \hat{b}_{l}^{k+1} = b_{l}^{k}$, and line 6 of Algorithm \ref{alg:Inertial}, we have

\begin{align*}
&\| \nabla_{b_{l}}\phi(a_{l-1}^{k+1},W_{l}^{k+1},z_{l}^{k+1},b_{l}^{k+1}) -\nabla_{{b}_{l}}\phi(\overline{a}_{l-1}^{k+1},\overline{W}_{l}^{k+1},\overline{z}_{l}^{k+1},\hat{b}_{l}^{k+1}) -\rho(b_{l}^{k+1}-\tilde{b}_{l}^{k+1}) \|       \\
&\leq   M\left(   \|a_{l-1}^{k+1} - \overline{a}_{l-1}^{k+1} \| + \|W_{l}^{k+1} - \overline{W}_{l}^{k+1} \| + \|z_{l}^{k+1} - \overline{z}_{l}^{k+1} \| + \|b_{l}^{k+1} - \hat{b}_{l}^{k+1} \|   \right) + \rho \| b_l^{k+1}- \tilde{b}_l^{k+1} \|    \\
& \leq   M\left(  \|b_{l}^{k+1} - b_{l}^{k} \|  \right) +  \rho \| b_l^{k+1}- b_l^{k} \| .      \quad  \gray{ \text{( Algorithm~\ref{alg:Inertial} )}  }
\end{align*}

According to the optimality condition of \eqref{eq:argminb}, there exists $g_{6,l}^{k+1} \in \partial_{b_l^{k+1}} F$ such that 
\begin{align}
\label{eq:subgradientb2}
\|g_{6,l}^{k+1}\| \leq   \left( M+\rho \right) \|b_l^{k+1} -  b_l^{k}  \|  .
\end{align}
Thus, combining \eqref{eq:subgradientb1} and \eqref{eq:subgradientb2}, we obtain $g_6^{k+1}=g_{6,1}^{k+1}\times g_{6,2}^{k+1}\times\cdots\times g_{6,L}^{k+1}\in\partial_{\mathbf{b}^{k+1}}F$ such that
\begin{align}
\label{eq:proofsubgradientb_l}
\|g_6^{k+1}\| \leqslant C_6\Big(\|\mathbf{b}^{k+1}-\mathbf{b}^k\|+ |\mathbf{b}^{k}-\mathbf{b}^{k-1}\| + \|\mathbf{W}^{k+1}-\mathbf{W}^k\|  + \|\mathbf{z}^{k+1}-\mathbf{z}^k\| + \|\mathbf{a}^{k+1}-\mathbf{a}^k\|\Big)    .
\end{align}

(2) For the parameter $z$, we divide it into two parts in the following proof. \\

(2.1) For $z_l^{k+1} (l<L)$, according to Algorithm \ref{alg:Inertial} and Algorithm \ref{alg:our}, and $z_l^{k+1}\leftarrow\tilde{z}_l^{k+1}-\nabla_{{z}_l}\phi \left(\overline{a}_{l-1}^{k+1},\overline{W}_{l}^{k+1},\hat{z}_{l}^{k+1},\overline{b}_{l}^{k+1}\right) /\rho$, we obtain $\nabla_{{z}_{l}}\phi = -\rho(z_{l}^{k+1}-\tilde{z}_{l}^{k+1})$. Thus, we have
\begin{align*}
\partial_{z_{l}^{k+1}}F&=\nabla_{z_{l}}\phi\left(a_{l-1}^{k+1},W_{l}^{k+1},z_{l}^{k+1},b_{l}^{k+1}\right)\\
&=\nabla_{z_{l}}\phi\left(a_{l-1}^{k+1},W_{l}^{k+1},z_{l}^{k+1},b_{l}^{k+1}\right)-\nabla_{{z}_{l}}\phi\left(\overline{a}_{l-1}^{k+1},\overline{W}_{l}^{k+1},\hat{z}_{l}^{k+1},\overline{b}_{l}^{k+1}\right) -\rho(z_{l}^{k+1}- \tilde{z}_{l}^{k+1})   .
\end{align*}

On one hand, using the above formula, the triangle inequality, and Assumption \ref{assu:2}, we obtain:
\begin{align*}
&\| \nabla_{z_{l}}\phi\left(a_{l-1}^{k+1},W_{l}^{k+1},z_{l}^{k+1},b_{l}^{k+1}\right)-\nabla_{{z}_{l}}\phi\left(\overline{a}_{l-1}^{k+1},\overline{W}_{l}^{k+1},\hat{z}_{l}^{k+1},\overline{b}_{l}^{k+1}\right) -\rho(z_{l}^{k+1}- \tilde{z}_{l}^{k+1})     \|       \\
&\leq \| \nabla_{z_{l}}\phi\left(a_{l-1}^{k+1},W_{l}^{k+1},z_{l}^{k+1},b_{l}^{k+1}\right)-\nabla_{{z}_{l}}\phi\left(\overline{a}_{l-1}^{k+1},\overline{W}_{l}^{k+1},\hat{z}_{l}^{k+1},\overline{b}_{l}^{k+1}\right)\| + \rho \|(z_{l}^{k+1}- \tilde{z}_{l}^{k+1})     \|   \\
&\leq  M\left(   \|a_{l-1}^{k+1} - \overline{a}_{l-1}^{k+1} \| + \|W_{l}^{k+1} - \overline{W}_{l}^{k+1} \| + \|z_{l}^{k+1} - \hat{z}_{l}^{k+1} \| + \|b_{l}^{k+1} - \overline{b}_{l}^{k+1} \|   \right)  + \rho \|(z_{l}^{k+1}- \tilde{z}_{l}^{k+1})     \|    \\
& \leq   M\Big(  \|a_{l-1}^{k+1} -  \left(  a_{l-1}^{k+1}+p_{3}\left(a_{l-1}^{k+1}-a_{l-1}^{k}\right) \right)  \| + \|W_{l}^{k+1} - \left  (W_{l}^{k+1}+p_{3}\left(W_{l}^{k+1}-W_{l}^{k}\right)  \right)   \|     \\
& + \|z_{l}^{k+1} - \left(   z_{l}^{k}+p_{2}\left(z_{l}^{k}-z_{l}^{k-1}\right)  \right) \| + \|b_{l}^{k+1} - \left(  b_{l}^{k+1}+p_{3}\left(b_{l}^{k+1}-b_{l}^{k}\right)  \right) \|  \Big)      \\
&+  \rho \| z_l^{k+1}- \left(  z_{l}^{k}+p_{1}\left(z_{l}^{k}-z_{l}^{k-1}\right)  \right) \|   \gray{  \quad \text{( Algorithm~\ref{alg:Inertial} )}  } \\
& \leq  M\Big(  \| a_{l-1}^{k+1}-a_{l-1}^{k}  \| + \| W_{l}^{k+1}-W_{l}^{k}  \|  + \| z_{l}^{k+1} - z_{l}^{k} \|  + \| z_{l}^{k}-z_{l}^{k-1} \| + \| b_{l}^{k+1}-b_{l}^{k} \|  \Big)      \\
&+  \rho \| z_l^{k+1}-  z_{l}^{k} \| +  \rho \|  z_{l}^{k}-z_{l}^{k-1} \| .    \gray{  \text{(Triangle Inequality)}   }   
\end{align*}

Otherwise, if the objective function $F(\cdot)$ increases after updating the parameter $z_l$, then according to Assumption \ref{assu:2}, $\tilde{z}_{l}^{k+1} = {z}_{l}^{k}, \hat{z}_{l}^{k+1} = {z}_{l}^{k}$, and line 6 of Algorithm \ref{alg:Inertial}, we obtain
\begin{align*}
&\| \nabla_{z_{l}}\phi\left(a_{l-1}^{k+1},W_{l}^{k+1},z_{l}^{k+1},b_{l}^{k+1}\right)-\nabla_{{z}_{l}}\phi\left(\overline{a}_{l-1}^{k+1},\overline{W}_{l}^{k+1},\hat{z}_{l}^{k+1},\overline{b}_{l}^{k+1}\right) -\rho(z_{l}^{k+1}- \tilde{z}_{l}^{k+1})     \|       \\
&\leq \| \nabla_{z_{l}}\phi\left(a_{l-1}^{k+1},W_{l}^{k+1},z_{l}^{k+1},b_{l}^{k+1}\right)-\nabla_{{z}_{l}}\phi\left(\overline{a}_{l-1}^{k+1},\overline{W}_{l}^{k+1},\hat{z}_{l}^{k+1},\overline{b}_{l}^{k+1}\right)\| + \rho \|(z_{l}^{k+1}- \tilde{z}_{l}^{k+1})     \|   \\
&\leq  M\left(   \|a_{l-1}^{k+1} - \overline{a}_{l-1}^{k+1} \| + \|W_{l}^{k+1} - \overline{W}_{l}^{k+1} \| + \|z_{l}^{k+1} - \hat{z}_{l}^{k+1} \| + \|b_{l}^{k+1} - \overline{b}_{l}^{k+1} \|   \right)  + \rho \|(z_{l}^{k+1}- \tilde{z}_{l}^{k+1})     \|    \\
& \leq   M \left( \|z_{l}^{k+1} - z_{l}^{k} \|   \right)  +  \rho \| z_l^{k+1}-  z_{l}^{k} \|.
\end{align*}

(2.2) For $z_L^{k+1}$, we have 
\begin{align*}
 & \partial_{z_{L}^{k+1}}F=\nabla_{z_{L}}\phi\left(a_{L-1}^{k+1},W_{L}^{k+1},z_{L}^{k+1},b_{L}^{k+1}\right)+\partial R\left(z_{L}^{k+1};y\right) \\
 & =\nabla_{z_{L}}\phi\left(a_{L-1}^{k+1},W_{L}^{k+1},z_{L}^{k+1},b_{L}^{k+1}\right)+\partial R\big(z_{L}^{k+1};y\big) +\nabla_{{z}_L}\phi\Big(\overline{a}_{L-1}^{k+1},\overline{W}_L^{k+1},\hat{z}_L^{k+1},\overline{b}_{L}^{k+1}\Big) +\rho\Big(z_{L}^{k+1}-\tilde{z}_{L}^{k+1}\Big) \\
 &-\nabla_{{z}_{L}}\phi\Big(\overline{a}_{L-1}^{k+1},\overline{W}_{L}^{k+1},\hat{z}_{L}^{k+1},\overline{b}_{L}^{k+1}\Big)-\rho\Big(z_{L}^{k+1}-\tilde{z}_{L}^{k+1}\Big)  .
 \end{align*}
According to the optimality condition of \eqref{eq:zL_argmin}, we have $0\in\partial R(z_L^{k+1};y)+\nabla_{{z}_L}\phi\left(\overline{a}_{L-1}^{k+1},\overline{W}_L^{k+1},\hat{z}_L^{k+1},\overline{b}_{L}^{k+1}\right) + \rho (z_L^{k+1}-\tilde{z}_L^{k+1})$. Therefore, we analyze the following formula. On the one hand, by applying the triangle inequality and Assumption \ref{assu:2}, we obtain
 \begin{align*}
 & \| \nabla_{z_L}\phi\left(a_{L-1}^{k+1},W_L^{k+1},z_L^{k+1},b_{L}^{k+1}\right)  -\nabla_{{z}_{L}}\phi\Big(\overline{a}_{L-1}^{k+1},\overline{W}_{L}^{k+1},\hat{z}_{L}^{k+1},\overline{b}_{L}^{k+1}\Big)-\rho\Big(z_{L}^{k+1}-\tilde{z}_{L}^{k+1}\Big)  \| \\
 &\leq  \| \nabla_{z_L}\phi\left(a_{L-1}^{k+1},W_L^{k+1},z_L^{k+1},b_{L}^{k+1}\right)  -\nabla_{{z}_{L}}\phi\Big(\overline{a}_{L-1}^{k+1},\overline{W}_{L}^{k+1},\hat{z}_{L}^{k+1},\overline{b}_{L}^{k+1}\Big) \| + \rho \| \Big(z_{L}^{k+1}-\tilde{z}_{L}^{k+1}\Big)  \|   \\
 & \leq   M\left(   \|a_{L-1}^{k+1} - \overline{a}_{L-1}^{k+1} \| + \|W_{L}^{k+1} - \overline{W}_{L}^{k+1} \| + \|z_{L}^{k+1} - \hat{z}_{L}^{k+1} \| + \|b_{L}^{k+1} - \overline{b}_{L}^{k+1} \|   \right) + \rho \| z_L^{k+1}- \tilde{z}_L^{k+1} \|          \\
 & \leq   M\Big(  \|a_{L-1}^{k+1} -  \left(  a_{L-1}^{k+1}+p_{3}\left(a_{L-1}^{k+1}-a_{L-1}^{k}\right) \right)  \| + \|W_{L}^{k+1} - \left  (W_{L}^{k+1}+p_{3}\left(W_{L}^{k+1}-W_{L}^{k}\right)  \right)   \|     \\
 & + \|z_{L}^{k+1} - \left(   z_{L}^{k}+p_{2}\left(z_{L}^{k}-z_{L}^{k-1}\right)  \right) \| + \|b_{L}^{k+1} - \left(  b_{L}^{k+1}+p_{3}\left(b_{L}^{k+1}-b_{L}^{k}\right)  \right) \|  \Big)      \\
 &+  \rho \| z_L^{k+1}- \left(  z_{L}^{k}+p_{1}\left(z_{L}^{k}-z_{L}^{k-1}\right)  \right) \|    \gray{  \quad \text{(Algorithm~\ref{alg:Inertial})}  } \\
 & \leq  M\Big(  \| a_{L-1}^{k+1}-a_{L-1}^{k}  \| + \| W_{L}^{k+1}-W_{L}^{k}  \|  + \| z_{L}^{k+1} - z_{L}^{k} \|  + \| z_{L}^{k}-z_{L}^{k-1} \| + \| b_{L}^{k+1}-b_{L}^{k} \|  \Big)      \\
 &+  \rho \| z_L^{k+1}-  z_{L}^{k} \| +  \rho \|  z_{L}^{k}-z_{L}^{k-1} \|  .   \gray{  \text{(Triangle Inequality)}   }
\end{align*}

Otherwise, if the objective function $F(\cdot)$ increases after updating parameter $z_L$, similar to $z_l$, then by applying line 6 of Algorithm \ref{alg:Inertial} and using $\tilde{z}_{L}^{k+1} = {z}_{L}^{k}, \hat{z}_{L}^{k+1} = {z}_{L}^{k}$, we obtain
 \begin{align*}
 & \| \nabla_{z_L}\phi\left(a_{L-1}^{k+1},W_L^{k+1},z_L^{k+1},b_{L}^{k+1}\right)  -\nabla_{{z}_{L}}\phi\Big(\overline{a}_{L-1}^{k+1},\overline{W}_{L}^{k+1},\hat{z}_{L}^{k+1},\overline{b}_{L}^{k+1}\Big)-\rho\Big(z_{L}^{k+1}-\tilde{z}_{L}^{k+1}\Big)  \| \\
 &\leq  \| \nabla_{z_L}\phi\left(a_{L-1}^{k+1},W_L^{k+1},z_L^{k+1},b_{L}^{k+1}\right)  -\nabla_{{z}_{L}}\phi\Big(\overline{a}_{L-1}^{k+1},\overline{W}_{L}^{k+1},\hat{z}_{L}^{k+1},\overline{b}_{L}^{k+1}\Big) \| + \rho \| \Big(z_{L}^{k+1}-\tilde{z}_{L}^{k+1}\Big)  \|   \\
 & \leq   M\left(   \|a_{L-1}^{k+1} - \overline{a}_{L-1}^{k+1} \| + \|W_{L}^{k+1} - \overline{W}_{L}^{k+1} \| + \|z_{L}^{k+1} - \hat{z}_{L}^{k+1} \| + \|b_{L}^{k+1} - \overline{b}_{L}^{k+1} \|   \right) + \rho \| z_L^{k+1}- \tilde{z}_L^{k+1} \|        \\
 & \leq   M\left(  \|z_{L}^{k+1} - z_{L}^{k}  \|  \right)  +  \rho \| z_L^{k+1} - z_{L}^{k}  \| .
 \end{align*}
 
In conclusion,  combining the two conditions above for $z_l ~and~ z_L$, there exists $g_{3,l}^{k+1} \in \partial_{z_{l}^{k+1}} F$ such that $g_3^{k+1}=g_{3,1}^{k+1}\times g_{3,2}^{k+1}\times\cdots\times g_{3,L}^{k+1} \in \partial_{\mathbf{z}^{k+1}}F$, and the coefficient of the corresponding subgradient is $C_3 = \max\left( M , M + \rho \right)$. Thus there exists $g_{3}^{k+1}$ such that
\begin{align}
\label{eq:proofsubgradientz_l_L}
\|g_3^{k+1}\| \leqslant C_3\Big(  \|\mathbf{a}^{k+1}-\mathbf{a}^k\| +  \|\mathbf{W}^{k+1}-\mathbf{W}^k\|  + \|\mathbf{z}^{k+1}-\mathbf{z}^k\|  + \|\mathbf{z}^{k}-\mathbf{z}^{k-1}\| + \|\mathbf{b}^{k+1}-\mathbf{b}^k\|  \Big)   .
\end{align}

(3) For $a_l^{k+1}$, according to $a_l^{k+1}\leftarrow\tilde{a}_l^{k+1}-\nabla_{{a}_l}\phi\left(\hat{a}_l^{k+1},\overline{W}_{l+1}^k,\overline{z}_{l+1}^k,\overline{b}_{l+1}^{k}\right)/\tau_l^{k+1}$, we obtain $\nabla_{{a}_l}\phi = -\tau_l^{k+1}\left(a_l^{k+1}-\tilde{a}_l^{k+1}\right) $, and thus
\begin{align*}
 \partial_{a_{l}^{k+1}}F & = \nabla_{a_{l}}\phi\left(a_{l}^{k+1},W_{l+1}^{k+1},z_{l+1}^{k+1},b_{l+1}^{k+1}\right) \\
 & =\nabla_{a_l}\phi\left(a_l^{k+1},W_{l+1}^{k+1},z_{l+1}^{k+1},b_{l+1}^{k+1}\right)-\nabla_{{a}_l}\phi\left(\hat{a}_l^{k+1},\overline{W}_{l+1}^k,\overline{z}_{l+1}^k,\overline{b}_{l+1}^{k}\right) -\tau_l^{k+1}\left(a_l^{k+1}-\tilde{a}_l^{k+1}\right)  .
\end{align*}
By applying the triangle inequality, we obtain
\begin{align*}
 &\|  \nabla_{a_l}\phi\left(a_l^{k+1},W_{l+1}^{k+1},z_{l+1}^{k+1},b_{l+1}^{k+1}\right)-\nabla_{{a}_l}\phi\left(\hat{a}_l^{k+1},\overline{W}_{l+1}^k,\overline{z}_{l+1}^k,\overline{b}_{l+1}^{k}\right) -\tau_l^{k+1}\left(a_l^{k+1}-\tilde{a}_l^{k+1}\right)  \|      \\
 & \leq  \|  \nabla_{a_l}\phi\left(a_l^{k+1},W_{l+1}^{k+1},z_{l+1}^{k+1},b_{l+1}^{k+1}\right)-\nabla_{{a}_l}\phi\left(\hat{a}_l^{k+1},\overline{W}_{l+1}^k,\overline{z}_{l+1}^k,\overline{b}_{l+1}^{k}\right) \|  + \tau_l^{k+1} \| 
 \left(a_l^{k+1}-\tilde{a}_l^{k+1}\right)  \|  ,
\end{align*}
 so according to Assumption \ref{assu:2}, we can acquire
\begin{align*}
 &\|\partial_{a_{l}^{k+1}}F\| \leq    \|  \nabla_{a_l}\phi\left(a_l^{k+1},W_{l+1}^{k+1},z_{l+1}^{k+1},b_{l+1}^{k+1}\right)-\nabla_{{a}_l}\phi\left(\hat{a}_l^{k+1},\overline{W}_{l+1}^k,\overline{z}_{l+1}^k,\overline{b}_{l+1}^{k}\right) \|  + \tau_l^{k+1} \| a_l^{k+1}-\tilde{a}_l^{k+1}  \|    \\
 & \leq   M\left(   \|a_{l}^{k+1} - \hat{a}_{l}^{k+1} \| + \|W_{l+1}^{k+1} - \overline{W}_{l+1}^{k} \| + \|z_{l+1}^{k+1} - \overline{z}_{l+1}^{k} \| + \|b_{l+1}^{k+1} - \overline{b}_{l+1}^{k} \|   \right) + \tau_l^{k+1}  \| a_{l}^{k+1}- \tilde{a}_{l}^{k+1} \|        .        
\end{align*}

On the one hand, according to Algorithm \ref{alg:Inertial}, we have
\begin{align*}
 & \|\partial_{a_{l}^{k+1}}F\| \leq  M\Big(  \|a_{l}^{k+1} -  \left(  a_{l}^{k}+p_{2}\left(a_{l}^{k}-a_{l}^{k-1}\right) \right)  \| + \|W_{l+1}^{k+1} - \left  (W_{l+1}^{k}+p_{3}\left(W_{l+1}^{k}-W_{l+1}^{k-1}\right)  \right)   \|     \\
 & + \|z_{l+1}^{k+1} - \left(   z_{l+1}^{k}+p_{3}\left(z_{l+1}^{k}-z_{l+1}^{k-1}\right)  \right) \| + \|b_{l+1}^{k+1} - \left(  b_{l+1}^{k}+p_{3}\left(b_{l+1}^{k}-b_{l+1}^{k-1}\right)  \right) \|  \Big)       \\  
 & + \tau_l^{k+1}  \| a_{l}^{k+1}-  \left(  a_{l}^{k}+p_{1}\left(a_{l}^{k}-a_{l}^{k-1}\right) \right)  \|     \\
 &\leq   M\Big(  \|a_{l}^{k+1} - a_{l}^{k}  \| + \| a_{l}^{k}-a_{l}^{k-1} \|  + \|W_{l+1}^{k+1} - W_{l+1}^{k} \|  + \| W_{l+1}^{k}-W_{l+1}^{k-1}  \|   + \|z_{l+1}^{k+1} -  z_{l+1}^{k} \|  \\ 
 &+ \| z_{l+1}^{k}-z_{l+1}^{k-1} \|  + \|b_{l+1}^{k+1} -  b_{l+1}^{k} \|  + \| b_{l+1}^{k}-b_{l+1}^{k-1} \|   \Big)  + \tau_l^{k+1}  \| a_{l}^{k+1}- a_{l}^{k} \|  + \tau_l^{k+1}  \| a_{l}^{k}-a_{l}^{k-1}  \|   ,
\end{align*}
where the last inequality follows from the triangle inequality and $0 \leq p_1 < 1$, $0 \leq p_2 < 1$, and $0 \leq p_3 < 1$. Therefore, there exists $g_{7,l}^{k+1} \in \partial_{a_l^{k+1}} F$ such that
\begin{align*}
& \| g_{7,l}^{k+1} \|  \leq   M\Big(  \|W_{l+1}^{k+1} - W_{l+1}^{k} \|  + \| W_{l+1}^{k}-W_{l+1}^{k-1}  \|   + \|z_{l+1}^{k+1} -  z_{l+1}^{k} \| + \| z_{l+1}^{k}-z_{l+1}^{k-1} \|  + \|b_{l+1}^{k+1} -  b_{l+1}^{k} \|  \\ 
& + \| b_{l+1}^{k}-b_{l+1}^{k-1} \|   \Big)  + \left( M +\tau_l^{k+1} \right) \left(\| a_{l}^{k+1}- a_{l}^{k} \|  + \tau_l^{k+1}  \| a_{l}^{k}-a_{l}^{k-1}  \| \right)    .
\end{align*}

Otherwise, if the objective function $F(\cdot)$ increases after updating the parameter $a$, according to Assumption \ref{assu:2} and line 6 of Algorithm \ref{alg:Inertial}, we have
\begin{align*}
 &\|\partial_{a_{l}^{k+1}}F\| \leq    \|  \nabla_{a_l}\phi\left(a_l^{k+1},W_{l+1}^{k+1},z_{l+1}^{k+1},b_{l+1}^{k+1}\right)-\nabla_{{a}_l}\phi\left(\hat{a}_l^{k+1},\overline{W}_{l+1}^k,\overline{z}_{l+1}^k,\overline{b}_{l+1}^{k}\right) \|  + \tau_l^{k+1} \| a_l^{k+1}-\tilde{a}_l^{k+1}  \|    \\
 & \leq   M\left(   \|a_{l}^{k+1} - \hat{a}_{l}^{k+1} \| + \|W_{l+1}^{k+1} - \overline{W}_{l+1}^{k} \| + \|z_{l+1}^{k+1} - \overline{z}_{l+1}^{k} \| + \|b_{l+1}^{k+1} - \overline{b}_{l+1}^{k} \|   \right) + \tau_l^{k+1}  \| a_{l}^{k+1}- \tilde{a}_{l}^{k+1} \|                     \\
 & \leq  M\Big(  \|a_{l}^{k+1} -  a_{l}^{k}  \| + \| W_{l+1}^{k+1} - W_{l+1}^{k}    \|  + \|z_{l+1}^{k+1} -  z_{l+1}^{k}  \| + \|b_{l+1}^{k+1} -   b_{l+1}^{k}  \|  \Big)  + \tau_l^{k+1}  \| a_{l}^{k+1}- a_{l}^{k}  \| .
\end{align*}

Therefore, there exists $g_{7,l}^{k+1} \in \partial_{a_{l}^{k+1}} F$ such that $C_7 = \max\left( M , M +\tau_{1}^{k+1}, M +\tau_{2}^{k+1},\cdots,  M +\tau_{L}^{k+1}\right)$ and $g_7^{k+1}=g_{7,1}^{k+1}\times g_{7,2}^{k+1}\times\cdots\times g_{7,L}^{k+1} \in \partial_{\mathbf{a}^{k+1}}F$. Combining the above two conditions, we have
\begin{align}
\label{eq:proofsubgradienta_l}
\left\|g_{7}^{k+1}\right\| & \leq C_{7} \Big(\|\mathbf{W}^{k+1}-\mathbf{W}^k\|+\|\mathbf{z}^{k+1}-\mathbf{z}^k\|+ \|\mathbf{a}^{k+1}-\mathbf{a}^k\| +\|\mathbf{b}^{k+1}-\mathbf{b}^k\|+\|\mathbf{W}^k-\mathbf{W}^{k-1}\|  \nonumber \\
& +\|\mathbf{b}^k-\mathbf{b}^{k-1}\| +\|\mathbf{z}^k-\mathbf{z}^{k-1}\|   +\|\mathbf{a}^k-\mathbf{a}^{k-1}\|   \Big)    .
\end{align}

In the end,  we combine \eqref{eq:proofsubgradientw_l}, \eqref{eq:proofsubgradientb_l}, \eqref{eq:proofsubgradientz_l_L}, and \eqref{eq:proofsubgradienta_l}, and $g^{k+1} \in \partial F(\mathbf{W}^{k+1},\mathbf{z}^{k+1},\mathbf{a}^{k+1},\mathbf{b}^{k+1}) = \{ \partial_{\mathbf{W}^{k+1}} F,\partial_{\mathbf{z}^{k+1}}F,\partial_{\mathbf{a}^{k+1}}F,\partial_{\mathbf{b}^{k+1}}F \}$ and $C_8$ = max($C_1,C_3,C_6,C_7$) such that

\begin{align}
\label{eq:totalsubgradient}    
\|g^{k+1}\| \leqslant & C_8 \Big(\|\mathbf{W}^{k+1}-\mathbf{W}^k\|+\|\mathbf{z}^{k+1}-\mathbf{z}^k\|+ \|\mathbf{a}^{k+1}-\mathbf{a}^k\| +\|\mathbf{b}^{k+1}-\mathbf{b}^k\|+\|\mathbf{W}^k-\mathbf{W}^{k-1}\|  \nonumber \\
& +\|\mathbf{b}^k-\mathbf{b}^{k-1}\| +\|\mathbf{z}^k-\mathbf{z}^{k-1}\|   +\| \mathbf{a}^k-\mathbf{a}^{k-1} \|  \Big).
\end{align}

In the final stage of the proof, we prove the convergence properties utilizing the KL property. Based on the proof of Lemma \ref{lemma:decrease}, as well as equations \eqref{eq:totalsubgradient} and \eqref{eq:FK-FK+1}, we have:
\begin{align*}
&F\Big(\mathbf{W}^{k},\mathbf{z}^{k},\mathbf{a}^{k},\mathbf{b}^{k}\Big)-F\Big(\mathbf{W}^{k+1},\mathbf{z}^{k+1},\mathbf{a}^{k+1},\mathbf{b}^{k+1}\Big)  \geq  C_5 \Big(\|\mathbf{W}^{k+1}-\mathbf{W}^k\|^2+\|\mathbf{z}^{k+1}-\mathbf{z}^k\|^2  + \|\mathbf{b}^{k+1}-\mathbf{b}^k\|^2+\|\mathbf{a}^{k+1}-\mathbf{a}^k\|^2  \Big),
\end{align*}
and
\begin{align*}
&F\Big(\mathbf{W}^{k-1},\mathbf{z}^{k-1},\mathbf{a}^{k-1},\mathbf{b}^{k-1}\Big)-F\Big(\mathbf{W}^{k},\mathbf{z}^{k},\mathbf{a}^{k},\mathbf{b}^{k}\Big)  \geq  C_5 \Big(\|\mathbf{W}^{k}-\mathbf{W}^{k-1}\|^2+\|\mathbf{z}^{k}-\mathbf{z}^{k-1}\|^2              +\|\mathbf{b}^{k}-\mathbf{b}^{k-1}\|^2+\|\mathbf{a}^{k}-\mathbf{a}^{k-1}\|^2  \Big)     .
\end{align*}
Summing the two above inequalities, we have
\begin{align*}
& F\Big(\mathbf{W}^{k-1},\mathbf{z}^{k-1},\mathbf{a}^{k-1},\mathbf{b}^{k-1}\Big)-F\Big(\mathbf{W}^{k+1},\mathbf{z}^{k+1},\mathbf{a}^{k+1},\mathbf{b}^{k+1}\Big)  \geq   C_5 \Big(  \|\mathbf{W}^{k}-\mathbf{W}^{k-1}\|^2+\|\mathbf{z}^{k}-\mathbf{z}^{k-1}\|^2             \\
&+\|\mathbf{b}^{k}-\mathbf{b}^{k-1}\|^2+\|\mathbf{a}^{k}-\mathbf{a}^{k-1}\|^2   +  \|\mathbf{W}^{k+1}-\mathbf{W}^k\|^2+\|\mathbf{z}^{k+1}-\mathbf{z}^k\|^2  +  \|\mathbf{b}^{k+1}-\mathbf{b}^k\|^2+\|\mathbf{a}^{k+1}-\mathbf{a}^k\|^2      \Big)   .
\end{align*}

In summary, based on the KL property, we have
\begin{align*}
&1 \leqslant \|g^{k+1}\|/ \left(\mu\sqrt{F\left(\mathbf{W}^{k+1},\mathbf{z}^{k+1},\mathbf{a}^{k+1},\mathbf{b}^{k+1}\right)-F^*}\right)          \\ 
&\leqslant C_8 \Big(\|\mathbf{W}^{k+1}-\mathbf{W}^k\|+\|\mathbf{z}^{k+1}-\mathbf{z}^k\|+ \|\mathbf{a}^{k+1}-\mathbf{a}^k\| +\|\mathbf{b}^{k+1}-\mathbf{b}^k\|+\|\mathbf{W}^k-\mathbf{W}^{k-1}\|  \nonumber \\
& +\|\mathbf{b}^k-\mathbf{b}^{k-1}\| +\|\mathbf{z}^k-\mathbf{z}^{k-1}\|   +\| \mathbf{a}^k-\mathbf{a}^{k-1} \| \Big)    /  \left(\mu\sqrt{F\left(\mathbf{W}^{k+1},\mathbf{z}^{k+1},\mathbf{a}^{k+1}\right)-F^*}\right)      \gray{   \text{(See \eqref{eq:totalsubgradient})}    }    \\
& \leqslant (C_{8})^{2} \Big(\|\mathbf{W}^{k+1}-\mathbf{W}^k\|+\|\mathbf{z}^{k+1}-\mathbf{z}^k\|+ \|\mathbf{a}^{k+1}-\mathbf{a}^k\| +\|\mathbf{b}^{k+1}-\mathbf{b}^k\|+\|\mathbf{W}^k-\mathbf{W}^{k-1}\|  \nonumber \\
& +\|\mathbf{b}^k-\mathbf{b}^{k-1}\| +\|\mathbf{z}^k-\mathbf{z}^{k-1}\|   +\| \mathbf{a}^k-\mathbf{a}^{k-1} \|  \Big)^{2}  /  \left(\mu^{2}\left(F\left(\mathbf{W}^{k+1},\mathbf{z}^{k+1},\mathbf{a}^{k+1},\mathbf{b}^{k+1}\right)-F^{*}\right)\right)     . 
\end{align*}
By applying the mean inequality, we get
\begin{align*}
& 1 \leqslant 8(C_{8})^{2} \Big(\|\mathbf{W}^{k+1}-\mathbf{W}^k\|^{2}+\|\mathbf{z}^{k+1}-\mathbf{z}^k\|^{2} + \|\mathbf{a}^{k+1}-\mathbf{a}^k\|^{2} +\|\mathbf{b}^{k+1}-\mathbf{b}^k\|^{2} +\|\mathbf{W}^k-\mathbf{W}^{k-1}\|^{2}  \nonumber \\
& +\|\mathbf{b}^k-\mathbf{b}^{k-1}\|^{2} +\|\mathbf{z}^k-\mathbf{z}^{k-1}\|^{2}   +\| \mathbf{a}^k-\mathbf{a}^{k-1} \|^{2}   \Big)  /  \left(\mu^{2}\left(F\left(\mathbf{W}^{k+1},\mathbf{z}^{k+1},\mathbf{a}^{k+1},\mathbf{b}^{k+1}\right)-F^{*}\right)\right)        \\ 
&\leq  8(C_{8})^{2}  \Big(  F\Big(\mathbf{W}^{k-1},\mathbf{z}^{k-1},\mathbf{a}^{k-1},\mathbf{b}^{k-1}\Big)-F\Big(\mathbf{W}^{k+1},\mathbf{z}^{k+1},\mathbf{a}^{k+1},\mathbf{b}^{k+1}\Big)  \Big) /   
 \\ 
& C_5 \mu^{2}\left(F\left(\mathbf{W}^{k+1},\mathbf{z}^{k+1},\mathbf{a}^{k+1},\mathbf{b}^{k+1}\right)-F^{*}\right)  .
\end{align*}
From the above formula, we can deduce that
$$\left(C_{5}\mu^{2}+8(C_{8})^{2}\right)\left(F\left(\mathbf{W}^{k+1},\mathbf{z}^{k+1},\mathbf{a}^{k+1},\mathbf{b}^{k+1}\right)-F^{*}\right)\leqslant8(C_{8})^{2}\left(F\left(\mathbf{W}^{k-1},\mathbf{z}^{k-1},\mathbf{a}^{k-1},\mathbf{b}^{k-1}\right)-F^{*}\right),$$
let $0<C_{9}=\frac{8(C_{8})^{2}}{C_{5}\mu^{2}+8(C_{8})^{2}}<1$. For any $\rho$, there exist $\epsilon > 0$, $k_1 \in \mathbb{N}$ and $0 < C_9 < 1$ such that for $k > k_1$, it holds that 
$$ \left(F\left(\mathbf{W}^{k+1},\mathbf{z}^{k+1},\mathbf{a}^{k+1},\mathbf{b}^{k+1}\right)-F^{*}\right)\leqslant C_9 \left(F\left(\mathbf{W}^{k-1},\mathbf{z}^{k-1},\mathbf{a}^{k-1},\mathbf{b}^{k-1}\right)-F^{*}\right).$$  
In other words, the linear convergence rate is proven.




\bibliographystyle{elsarticle-num} 






\end{document}